\documentclass[11pt]{article}
\usepackage[utf8]{inputenc}
\usepackage[left=2.5cm, right=2.5cm, top=2.5cm]{geometry}
\usepackage{natbib} % for citations
\usepackage{tikz} % for dags
\usepackage{setspace} % for spacing
\usepackage{amsmath}  % amsmath in combination with hyperref leads to weird spacing in equations, solution is to load amsmath before hyperref
\usepackage[hidelinks]{hyperref}  % generate links for citations
\usepackage{amssymb}
\usepackage{algorithm}
\usepackage{algpseudocode}
\usepackage{booktabs} % for pretty tables
\usepackage{tabularx} % for line breaks and lis ts within tables
\usepackage{pdflscape} % to rotate tables
\usepackage{mathtools} % to annotate parts of equations
\usepackage{listings}
\usepackage[column=O]{cellspace} % for appropriate whitespace in tables

\usepackage{multibib}  % to have separate references in appendix
\newcites{APX}{Appendix References}

\usepackage{enumitem}  % to remove the indent in itemize
\setitemize{noitemsep}

\usepackage{rotating}  % to rotate column labels
\usepackage{xparse}  % to rotate column labels
\NewDocumentCommand{\rot}{O{90} O{.005em} m}{\makebox[#2][l]{\rotatebox{#1}{#3}}}

\usepackage{caption}  % for captions, duh
\usepackage{subcaption}

\pgfdeclarelayer{bg}    % declare background layer
\pgfsetlayers{bg,main}  % set the order of the layers (main is the standard layer)

\usepackage[normalem]{ulem}
\useunder{\uline}{\ul}{}
\usepackage{ragged2e}
\usepackage{longtable}
\usepackage{authblk}
\usepackage[section]{placeins}  % to make figures/tables appear in the correct section

\xdefinecolor{ut-red}{RGB}{165,30,55}
\xdefinecolor{ut-gold}{RGB}{180,160,105}
\xdefinecolor{ut-dark}{RGB}{50,65,75}
\makeatletter
\g@addto@macro\normalsize{%
  \setlength\abovedisplayskip{8pt}
  \setlength\belowdisplayskip{8pt}
  \setlength\abovedisplayshortskip{2pt}
  \setlength\belowdisplayshortskip{2pt}
}
\makeatother

\usetikzlibrary{shapes,decorations,arrows,calc,arrows.meta,fit,positioning}
\tikzset{
    -Latex,auto,node distance =1 cm and 1 cm,semithick,
    state/.style ={ellipse, draw, minimum width = 0.7 cm},
    point/.style = {circle, draw, inner sep=0.04cm,fill,node contents={}},
    bidirected/.style={Latex-Latex,dashed},
    el/.style = {inner sep=2pt, align=left, sloped}
}
\title{Estimating Causal Effects with Double Machine Learning - A Method Evaluation}
\author[1]{Jonathan Fuhr}
\author[2]{Philipp Berens}
\author[1]{Dominik Papies}
\affil[1]{School of Business and Economics, University of Tübingen,  Tübingen, Germany}
\affil[2]{Hertie Institute for AI in Brain Health, University of Tübingen, Tübingen, Germany}
\date{Last edited: \today}

\begin{document}
\doublespacing

\maketitle

\begin{abstract}  
The estimation of causal effects with observational data continues to be a very active research area. In recent years, researchers have developed new frameworks which use machine learning to relax classical assumptions necessary for the estimation of causal effects. In this paper, we review one of the most prominent methods - ``double/debiased machine learning" (DML) - and empirically evaluate it by comparing its performance on simulated data relative to more traditional statistical methods, before applying it to real-world data. Our findings indicate that the application of a suitably flexible machine learning algorithm within DML improves the adjustment for various nonlinear confounding relationships. This advantage enables a departure from traditional functional form assumptions typically necessary in causal effect estimation. However, we demonstrate that the method continues to critically depend on standard assumptions about causal structure and identification. When estimating the effects of air pollution on housing prices in our application, we find that DML estimates are consistently larger than estimates of less flexible methods. From our overall results, we provide actionable recommendations for specific choices researchers must make when applying DML in practice.
\end{abstract}

\clearpage
\section{Introduction} \label{introduction}

In many scientific disciplines, researchers attempt to estimate causal effects \citep{imbens_causal_2015}, i.e., they ask how one variable of interest (the ``treatment") causally affects another variable (the ``outcome"). For example, labor economists care about the effect of education on wages \citep[e.g.,][]{card_causal_1999}, physicians want to know whether smoking causes lung cancer \citep[e.g.,][]{cornfield_smoking_2009}, and marketing managers want to know how a price change affects demand  \citep[e.g.,][]{bijmolt_new_2005}.
For many of these questions, researchers have to rely on observational data because experimental interventions may be infeasible, unethical, or simply too costly to obtain \citep[e.g.,][]{athey_state_2017}. However, without experimental variation, identifying and estimating causal effects is not possible without making assumptions. These assumptions can be strong and are generally not testable in observational studies \citep[e.g.,][]{pearl_causality_2009, imbens_causal_2015}. 
For example, a typical assumption is that all variables affecting both the treatment and the outcome variable (so-called ``confounders") are observed and adequately adjusted for \citep[e.g.,][]{imbens_nonparametric_2004}. 
Justifying the necessary assumptions is one of the biggest challenges in real-world applications of causal inference (e.g., \citealp[]{hernan_causal_2020}). As a consequence, researchers are interested in developing and applying methods that work under weaker or more plausible assumptions. 

Indeed, researchers have recently suggested that some of the necessary assumptions can be relaxed by relying on new causal inference frameworks based on \textit{machine learning}. Traditionally, (supervised) machine learning (ML) has established itself as a powerful tool for making \textit{predictions} in complex, nonlinear settings. ML methods are also capable of handling high-dimensional data, i.e., data where the number of variables or parameters may even exceed the number of observations \citep[see, e.g.,][]{hastie_elements_2009}. However, the goal of achieving high predictive accuracy is fundamentally different from the typical goal of causal inference, which is accurate parameter or effect estimation \citep{shmueli_explain_2010, mullainathan_machine_2017}. A consequence of these different goals is that directly using ML methods designed for prediction ``off-the-shelf" does not lead to parameter estimates that can be interpreted as causal effects \citep{athey_impact_2019}.  In other words, these methods do not usually provide unbiased estimates of causal effects, i.e., estimates that are on average equal to the true parameter value \citep{wooldridge_introductory_2012}. 
However, there may be a way to use ``machine learning in the service of causal inference" \citep{mullainathan_machine_2017}: The core tenet underlying this idea is that, in addition to causal assumptions, answering causal questions often requires an estimation process that we can sometimes divide into multiple prediction problems. For example, the first stage of instrumental variable estimation is a prediction task: we predict the treatment from the instrument(s) \citep{mullainathan_machine_2017}. Using data-driven ML for such predictive parts may facilitate a higher flexibility and may potentially allow for less restrictive assumptions in the estimation. 

In this paper, we focus on ``double/debiased machine learning" (DML) by \citet{chernozhukov_doubledebiased_2018}, which is arguably one of the most prominent examples of such a method using ML for causal inference.  The basic promise of DML is that we can use flexible ML methods to adequately adjust for observed confounding variables. With that, researchers can still obtain unbiased estimates, even in settings with potentially many confounders and complex functional forms. 
By using DML with a flexible ML method instead of specifying a parametric model, researchers may be able to relax assumptions about how (which variables, which functional forms) they adjust for observed confounding. This is important because flexibly adjusting for a large number of covariates can increase the plausibility of the assumption that all relevant confounding variation has been considered \citep{belloni_inference_2016}. 

It is important that applied researchers seeking to use DML fully appreciate all potential benefits, pitfalls, and assumptions of this method. In addition, when using DML, researchers must make a variety of choices and answer questions such as: ``%Do the required assumptions hold in the given application?%
Which variables should enter the estimation? Which ML algorithms should I use? Do these decisions depend on the sample size? How influential are these choices?" With these considerations and questions in mind, we see four important aspects currently missing from the literature: (1) an extensive review and discussion of DML, focused on an intuitive understanding and an assessment of the necessary assumptions, (2) an empirical evaluation of the method's performance, i.e., its ability to recover causal effects in a wide range of simulated settings mimicking causal problems encountered in applied fields, (3) specific guidance for the many choices researchers have to make when applying DML to their causal questions, and (4) an application of various implementations of DML to real-world data. To address these voids, we review and evaluate the method and compare it to more traditional statistical methods in both simulations and an application, from which we provide specific and actionable best-practice guidance to applied researchers.

To preview the results, a first finding is that the functional form of the confounding as well as the number of confounders strongly affect the suitability of specific ML algorithms for DML. More specifically, while many applications in the past have used lasso regression in the context of DML, we caution against its use in this context because DML using lasso without manual variable transformations produces biased estimates in the presence of nonlinear confounding. Second, the results suggest that the main advantage of DML with flexible ML methods is its ability to adjust for nonlinear confounding without knowing the underlying functional forms, rather than adjusting for a very large number of important confounders simultaneously. Third, we find that gradient boosting (XGBoost) performs very well across a broad range of settings in our analyses, which is why we recommend it as a  baseline or default method within DML. Further, the results also show that DML continues to critically hinge on researchers' input about causal structure and is no automatic remedy for unobserved confounding or bad controls \citep{cinelli_crash_2022}. 
Finally, to support researchers in their choice of a suitable ML algorithm, we present a simple metric that researchers can use to aid their selection of ML algorithms.

We structure the rest of the paper as follows. In Section \ref{sec:literature}, we review the current literature around DML. Section \ref{sec:methods} provides a mostly non-technical review of the method, focusing on the partially linear model. In Section \ref{sec:simulations}, we assess DML in a variety of simulation settings, where we compare its performance to more traditional methods and evaluate strengths and weaknesses. Section \ref{sec:application} applies DML to real-world data, which leads to a discussion of the plausibility of the results and the remaining assumptions. Section \ref{sec:furth_settings} briefly outlines the applicability of DML in settings beyond the partially linear model.   In Section \ref{sec:discussion}, we derive recommendations of best practice for applied researchers about when and how to apply this method, and finally conclude our discussion. 
\section{Literature review} \label{sec:literature}

We review the literature related to double/debiased machine learning \citep{chernozhukov_doubledebiased_2018} from three perspectives: (1) the history and purpose of DML, (2) a quantitative analysis of published applications of DML, and (3) method evaluations considering DML.

(1) First, one central purpose of DML is relaxing assumptions about model specification and functional forms of covariates. Even in cases where all confounding variables are observed, including them with a wrong functional form in a parametric regression model will lead to biased estimates of the treatment effect (e.g., \citealp{wooldridge_introductory_2012}, Chapter 9). Classical semi-parametric methods, which only make functional form assumptions about the treatment parameter, but not about the covariates, may alleviate this problem \citep{athey_impact_2019}. However, they tend to be slow to converge and require more observations compared to parametric methods \citep{powell_estimation_1994}. These methods are especially inadequate in high-dimensional settings, i.e., when the number of parameters to estimate is large relative to the sample size. On the one hand, this can occur through having more variables/features than observations ($p > n$), which is problematic for many traditional approaches in statistics 
(e.g., \citealp[]{james_introduction_2021}, Chapter 6). On the other hand, this can also occur when including various transformations and interactions of a relatively low-dimensional set of covariates to be more robust against model misspecification. Estimating parameters for each of these transformations could quickly lead to a high-dimensional setting \citep{belloni_high-dimensional_2014}. 
One example for a semi-parametric regression method is \citet{robinson_root-n-consistent_1988}, who models the relationship between covariates and both outcome and treatment using a kernel regression. DML builds on this method and provides a framework in which the kernel regressions can be replaced by modern ML methods \citep{chernozhukov_doubledebiased_2018}, which enables the application in high-dimensional settings. In a parallel development within biostatistics,  \citet{laan_targeted_2006} developed ``Targeted Maximum Likelihood Estimation" (TMLE), a semi-parametric estimation technique similar to DML that also allows for the use of ML methods. For a detailed review and comparison of TMLE and DML from a biostatistics perspective, see \citet{diaz_machine_2020}. 

A more direct predecessor of DML is the ``double selection" procedure \citep[][]{belloni_high-dimensional_2014}, which consists of three steps: First, it uses lasso regression as the ML method to select which covariates are predictive of the outcome; second, it uses lasso to select which covariates are predictive of the treatment; third, it uses OLS to regress the outcome on the treatment and the union of all covariates selected in the first two steps. This method can deliver unbiased estimates in situations where there are many potential confounding variables (potentially more than observations) from which only a few are important for adjustment (the ``sparsity assumption") \citep{belloni_high-dimensional_2014}. DML generalizes this idea and introduces a sample splitting procedure, which facilitates the application of many modern ML methods beyond lasso regression. 

\begin{figure}[ht]
    \centering
    \includegraphics[width=\textwidth]{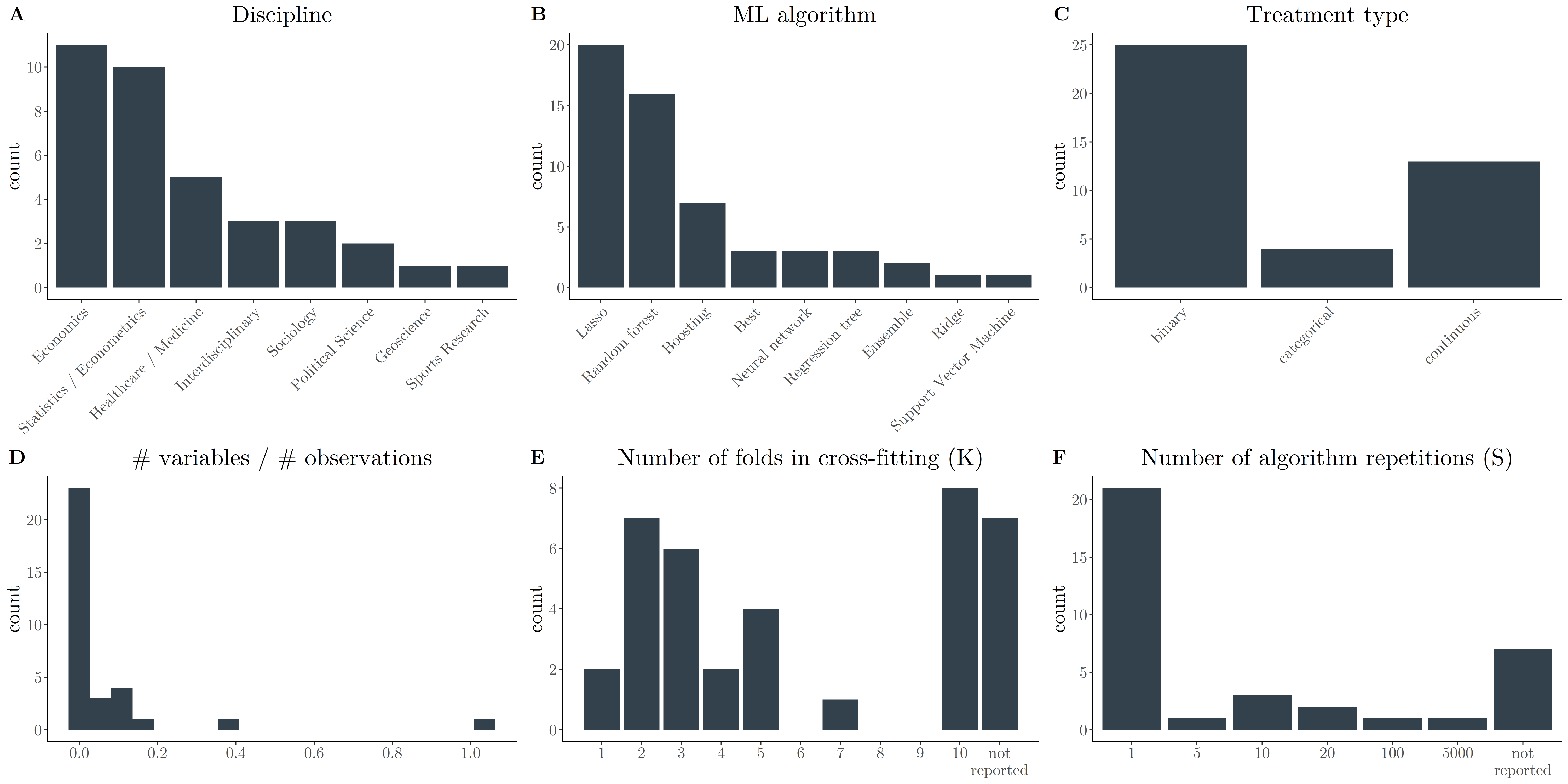}
    \caption{Overview of DML applications in the literature. \textbf{A} Discipline the application was published in. \textbf{B} Different ML algorithms used within DML. \textbf{C} Treatment type considered in application. \textbf{D} Dimensionality: ratio of the number of variables to the number of observations. \textbf{E} Number of folds the data is split into within DML. \textbf{F} Number of algorithm repetitions for increased robustness.}
    \label{fig:appl_lit}
\end{figure}

(2) As a second perspective, we review applications of DML published across disciplines. Overall, we have identified 46 published papers containing applications of DML to real-world data, of which 36 contained sufficient information about the implementation to be considered in this overview (see Appendix \ref{appx:lit_selection} for a list of these papers and a detailed description of our selection process). We summarize important characteristics of these applications (Figure \ref{fig:appl_lit}). Since \citet{chernozhukov_doubledebiased_2018} published the original paper in econometrics, a majority of subsequent applications occurred either directly in economics, or as part of further method developments within statistics or econometrics. However, the method has also received widespread attention in other quantitative disciplines such as healthcare/medicine, or sociology (Figure \ref{fig:appl_lit}A). 

Within DML, researchers most often used lasso and random forests for the predictive parts, followed by boosting methods (Figure \ref{fig:appl_lit}B). On average, an application considered 1.55 different ML methods. However, most papers (27/36, 75\%) used only one ML algorithm and do not assess the robustness of their estimates to different predictive methods. A likely reason for the dominance of lasso is the early implementation of DML in the statistical software Stata \citep{statacorp_stata_2019}, which uses lasso and which multiple papers explicitly mentioned. As we will discuss below, our results show that lasso without manual variable transformations tends to produce biased results in settings that involve nonlinear confounding.

A majority of the applications investigate effects of a binary treatment variable (Figure \ref{fig:appl_lit}C), but there are also many continuous treatments and a few categorical/multilevel treatment settings. 

One advantage of DML is its ability for valid inference in high-dimensional settings. However, there are arguably few real-world applications where the number of raw variables exceeds the number of observations. Hence, high-dimensionality in practice might rather be a result of flexibly modeling nonlinear relationships with many parameters. 
In the applications of DML in the literature, the distribution of the ratio of the number of variables to the number of observations supports this hypothesis (Figure \ref{fig:appl_lit}D). We define the number of variables as the raw covariates, not including transformations such as polynomials or interactions. Only one application is high-dimensional in raw covariates: \citet{chan_behind_2022} estimate the effect of technological intensity on support for the EU FDI screening mechanisms, using one observation for each of the 28 EU member states (n = 28) and adjusting for 29 variables. In all other applications, the number of raw covariates is significantly smaller than the number of observations. However, these settings can still benefit from high-dimensional methods if one includes many transformations of the raw covariates or if one uses flexible non-parametric ML methods like random forests or neural networks, which estimate many parameters in the training process to fit complex functional forms. 

Finally, when applying DML, researchers have to choose two specific parameters in the algorithm. First, $K$ is the number of folds into which the algorithm splits the data, using $K-1$ parts to train the ML model, and estimating the effects on the remaining part. Hence, $K = 1$ means no sample splitting, which does technically not fit the definition of DML in \citet{chernozhukov_doubledebiased_2018}. In their foundational paper, \citet{chernozhukov_doubledebiased_2018} introduce the method with two folds, but recommend four or five folds, because larger numbers of folds allow the ML methods to train on larger samples. In the applications we observed, choices of two and three folds are popular, while the most frequent number of folds is 10 (Figure \ref{fig:appl_lit}E). Again, this is likely due to the popular Stata implementation, which uses $K = 10$ as default.

The second parameter is the number of repetitions of the full DML algorithm in the applications. \citet{chernozhukov_doubledebiased_2018} state that the random sample splitting can have an effect on the estimates in finite samples and thus recommend repeating the algorithm $S$ times (they use $S = 100$) and reporting the median estimate across the repetitions. A majority of applications does not follow this advice (Figure \ref{fig:appl_lit}F), which we can partly explain by the Stata default ($S = 1$), but also by the $S$-fold increase in computation time. However, some publications (8/36, 22\%) use multiple repetitions. The largest number is 5000 repetitions in \citet{chan_behind_2022}, which is feasible (and maybe necessary) because of their small sample size of 28 observations. Our results below will show that in smaller samples, using a larger number of repetitions can increase the robustness of the estimates considerably.

To sum up, in most of the applications we considered, researchers do not explicitly establish why their choice of specific ML algorithms, number of folds and number of repetitions is appropriate for their application. Therefore, our paper aims to provide guidance about the impact of these choices and potential trade-offs.

(3) Other papers have assessed the performance of DML when comparing different ML-based causal inference methods, albeit with a focus different from ours: They typically compare one (or a few) specific implementation(s) of DML to other novel or established methods, whereas we focus mostly on how different implementations of DML can perform in a large variety of settings. Here, we review these publications that assess the accuracy of DML and other methods when estimating causal effects (Table \ref{tab:evals}). These evaluations come from different disciplines: three (\citet{mcconnell_estimating_2019, zivich_machine_2021, loiseau_external_2022}) from healthcare/medicine, two (\citet{yang_double_2020, gordon_close_2022}) from economics, and one (\citet{qiu_statistical_2022}) from geoscience. Most evaluations consider the average treatment effect of a binary treatment and assess the methods on simulated data. Exceptions are \citet{qiu_statistical_2022}, who work with a continuous treatment, and \citet{gordon_close_2022} as well as \citet{loiseau_external_2022}, who evaluate the methods by comparing their estimates to results from randomized controlled trials (RCTs). In our study, we also evaluate the methods on simulated data, but focus on settings with a continuous treatment variable (e.g., medication dosage, temperature, price, etc.).

In the first evaluation that included DML, \citet{mcconnell_estimating_2019} assess how different ML-based causal methods perform when estimating average treatment effects. They use random forests in DML and show that, while DML easily dominates traditional methods, other methods (e.g., Bayesian Causal Forests) are slightly more accurate in several cases. 
\citet{yang_double_2020} compare tree-based methods in DML for three different functional forms of confounding and find that gradient boosting in DML is superior to regression trees and random forests across their simulation settings. 
\citet{zivich_machine_2021} demonstrate the benefits of doubly robust methods and cross-fitting when estimating effects with ML. They find that methods like DML that use these techniques outperform singly-robust methods and methods without cross-fitting, but are computationally more expensive.
In contrast to the previous studies, \citet{gordon_close_2022} assess whether DML (using neural networks) can recover experimental estimates of advertising effectiveness at Facebook. While DML delivers more accurate estimates than a naive comparison and an alternative observational method (stratified propensity score matching), it is still unable to recover the benchmark from the RCTs. The authors explain that Facebook does not record all relevant targeting data, which violates the unconfoundedness assumption in their DML application.  
\citet{loiseau_external_2022} also assess multiple methods by comparing their estimated effects to RCT estimates, in addition to a simulation study. Their findings suggest using DML for larger sample sizes, but G-computation when only having access to smaller datasets. However, they only consider settings with few covariates and correctly specified models in their simulations, and use only relatively inflexible (regularized) linear methods for the predictive parts of DML. They call for further research on the performance benefits of using more flexible (e.g., tree-based) ML methods. 
Finally, \citet{qiu_statistical_2022} simulate data from a chemical transport model to assess the ability of various statistical approaches to adjust for meteorological variability when estimating effects of emission changes. In their setup, DML with random forests by far outperforms both traditional approaches and DML with lasso, but is not unbiased in all areas they consider. 

Our study adds four contributions to these evaluations: 
(1) we extensively introduce and review the DML framework with a focus on intuitive understanding, (2) we apply DML to a much wider range of simulated settings to assess boundary conditions for the method, (3) we compare a variety of predictive methods within the DML framework, and (4) we provide guidance on decisions researchers have to make when applying DML in their field.

\def\arraystretch{1.5}%  

\begin{landscape}
\begin{table}%[ht]
\tiny
\caption{Method evaluations considering DML}
\resizebox{\columnwidth}{!}{%
\begin{tabular}{>{\raggedright\arraybackslash}p{\dimexpr.08\textwidth}  p{\dimexpr.08\textwidth} p{\dimexpr.001\textwidth} p{\dimexpr.001\textwidth} p{\dimexpr.001\textwidth} p{\dimexpr.001\textwidth} p{\dimexpr.001\textwidth} p{\dimexpr.001\textwidth} p{\dimexpr.001\textwidth} p{\dimexpr.001\textwidth}  >{\raggedright\arraybackslash}p{\dimexpr.1\textwidth} >{\raggedright\arraybackslash}p{\dimexpr.12\textwidth} p{\dimexpr.2\textwidth} p{\dimexpr.3\textwidth}}
\toprule
& & \multicolumn{8}{c}{\textbf{ML methods} \vspace{.5em}}  &  &  & & \\
\textbf{Paper} & \textbf{Treatment type} & \multicolumn{1}{l}{\rot{\textbf{\hspace{-3ex}Linear regression}}} & \multicolumn{1}{l}{\rot{\textbf{\hspace{-3ex}Lasso/Ridge}}} & \multicolumn{1}{l}{\rot{\textbf{\hspace{-3ex}GAMs}}}  & \multicolumn{1}{l}{\rot{\textbf{\hspace{-3ex}Regression trees}}} & \multicolumn{1}{l}{\rot{\textbf{\hspace{-3ex}Random forests}}} & \multicolumn{1}{l}{\rot{\textbf{\hspace{-3ex}Boosting}}} & \multicolumn{1} {l}{\rot{\textbf{\hspace{-3ex}Neural networks}}} & \multicolumn{1} {l}{\rot{\textbf{\hspace{-3ex}Ensemble}}}  & \textbf{Alternative methods} & \textbf{Evaluation method} & \textbf{Parameters} & \textbf{Findings/recommendations}\\
\midrule
\citet{mcconnell_estimating_2019} & Binary &  &  &  &  & \checkmark  &  &   & 
& OLS, AIPW, TMLE, BART, ps-BART, BCF, GRF & Simulations & Confounding strength, Sample size (n), \# of confounders (p), \# of noise variables & DML slightly worse than BCF and ps-BART, but better than rest. DML is best if many covariates ($>$150).\\
\citet{yang_double_2020} & Binary &   &   & & \checkmark & \checkmark & \checkmark &  &
& - & Simulations & Various functional forms, \# of confounders (p), Sample size (n) & Gradient boosting in DML performs better than regression trees and random forests across DGPs. Bias increases in p, decreases in n.\\
\citet{zivich_machine_2021} & Binary &   &  &  &  &  &  &  &  \checkmark  
& G-computation, IPW, AIPW, TMLE, CF-TMLE & Simulations modeled after medical application & Provide different model specifications to methods (True, linear, ML), \# repetitions  & Doubly robust methods with cross-fitting and ML outperform singly-robust and not cross-fit ML methods, but need more computing resources. \\
\citet{gordon_close_2022} & Binary &  &  &  &  &  &  & \checkmark &   & Stratified PS-Matching (SPSM) & Comparison to RCTs from ad campaigns at Facebook & - & DML and SPSM unable to recover experimental benchmark. Both perform better than naive comparison. DML more accurate than SPSM. Reason: Not all relevant targeting data is logged and available for adjustment.\\
\citet{loiseau_external_2022} & Binary &  & \checkmark &  &  &  &  &   & &  PS-matching, IPTW, G-computation & Simulation and comparison to RCTs of diabetes medication & Sample size, Homogeneous vs heterogeneous treatment effect & DML has smallest bias, G-computation smallest MSE. DML improves with sample size. Recommend G-computation for n\textless 100, DML for n\textgreater 500.\\
\citet{qiu_statistical_2022} & Continuous &  & \checkmark &  &  & \checkmark &  &   &  & OLS, polynomial regression, cubic splines, GAMs & Simulations from a chemical model for air quality effects of emission changes & - & DML with random forests performs best, but still not ideal in all areas. Other methods mostly perform poorly.\\
\addlinespace
Our evaluation & Continuous & \checkmark & \checkmark & \checkmark &  & \checkmark & \checkmark & \checkmark  & &  OLS, Naive XGBoost & Simulations & Various functional forms; Confounding strength; \# of confounders (p); Sample size (n); Inclusion of noise variables; Inclusion of variables related to outcome; Inclusion of variables related to treatment; Violation of unconfoundedness; Inclusion of bad controls & XGBoost, neural networks and random forests in DML perform best across a variety of settings. Choose between algorithms based on predictive accuracy in first stage.  Decisions about causal structure and (parts of) variable selection not made automatically by DML but should be based on theory. Choose $K$ based on sample size, test different $S$ and choose number with stable results. \\
\bottomrule
\end{tabular}
}
\begin{tabularx}{\columnwidth}{X}\textit{Note: }OLS: Ordinary least squares, AIPW: Augmented inverse probability weighted estimator, TMLE: Targeted maximum likelihood estimator, BCF: Bayesian Causal Forests, BART: Bayesian additive regression trees, PS: Propensity score, GRF: Generalized random forests,  CF: cross-fit, GAMs: Generalized additive models\end{tabularx}
\label{tab:evals}
\end{table}
\end{landscape}

\section{Method review} \label{sec:methods}

\subsection{DML in the partially linear model}

In this section, we review double/debiased machine learning \citep{chernozhukov_doubledebiased_2018}, with the goal of providing an intuition about when, how and why the method works. One important setting in which researchers can apply DML to potentially relax assumptions is causal effect estimation in the presence of observed confounding (Figure \ref{fig:dag1}).  

\begin{figure}[ht]
\small
\centering
\begin{tikzpicture}
    \node (w) at (0,0) [label=left:$W$,point];
    \node (y) at (6,0) [label=right:$Y$,point];
    \node (xc) at (3,1.) [label=above:$\boldsymbol{X_c}$,point];

    \path (w) edge (y);
    \path (xc) edge node[above, el] {$g_0(\boldsymbol{X_c})$} (y);
    \path (xc) edge node[above, el] {$m_0(\boldsymbol{X_c})$} (w);
\end{tikzpicture}
\caption{\label{fig:dag1}Directed acyclic graph (DAG) for the assumed causal structure. $W$: treatment variable, $Y$: outcome variable, $\boldsymbol{X_c}$: observed confounding variables. The relationships between $\boldsymbol{X_c}$ and $W$ ($m_0()$), and $\boldsymbol{X_c}$ and $Y$ ($g_0()$), are potentially complex and nonlinear.}
\vspace{-2ex}
\end{figure}

In the assumed causal structure of Figure \ref{fig:dag1}, we want to estimate the causal effect of a treatment variable $W$ on an outcome variable $Y$. However, observed confounders $\boldsymbol{X_c}$ complicate the estimation by influencing both the treatment and the outcome. If we do not adequately adjust for such confounders, the estimated causal effect is biased (e.g., \citealp{hernan_causal_2020}, Chapter 7), i.e., different from the true value the estimator was supposed to provide \citep{wooldridge_introductory_2012}.
We can express the data-generating process (DGP) of this setting with a partially linear regression (PLR) model, where $V_y$ and $V_w$ are noise terms:
\begin{gather} 
Y = \beta W + g_0(\boldsymbol{X_c}) + V_y \label{eq:plry} \\
W = m_0(\boldsymbol{X_c}) + V_w  \label{eq:plrw}
\end{gather}
In the partially linear outcome model, we assume the treatment to have a linear, additive functional form, while the functional form of the confounders can be either linear or nonlinear. In this setting, the treatment has a constant, homogeneous effect on the outcome. We use the partially linear model for the introduction of DML, but DML is not limited to this setting. In case of a binary treatment, the treatment variable can arbitrarily interact with the confounders, which allows for the presence of heterogeneous treatment effects (see Section \ref{sec:furth_settings}).  

The assumptions encoded in Figure \ref{fig:dag1} as well as Equations \ref{eq:plry} and \ref{eq:plrw} imply that we can identify the causal effect of $W$ on $Y$ after adjusting for observed confounders $\boldsymbol{X_c}$ (\citealp{hernan_causal_2020}, Chapter 2). Different disciplines use different names for this assumption (with slightly different technical definitions): unconfoundedness, exogeneity, ignorability, selection on observables \citep{imbens_recent_2009}, conditional independence assumption \citep{angrist_mostly_2009}, exchangeability \citep{hernan_causal_2020}, or a case for backdoor adjustment \citep{pearl_causal_2016}. This assumption is made in many studies across disciplines and is therefore highly relevant for applied research \citep{imbens_recent_2009}.

However, even if this assumption holds, two challenges appear. First, the functional relationships of the confounders with the treatment ($m_0(\boldsymbol{X_c})$) and the outcome ($g_0(\boldsymbol{X_c})$) may be nonlinear and complex.
Secondly, there could be a large number of potential confounders in $\boldsymbol{X_c}$, such that it is not obvious which covariates one should include in the estimation process. Such a high-dimensional setting can either arise naturally through many observed variables (e.g., covariates generated from text, images, or genes), or when including transformations of observed variables (e.g., polynomials, interactions, etc.) to allow for potential nonlinearities in the model \citep{belloni_high-dimensional_2014}. Traditionally, researchers would make additional assumptions to address these challenges. For example, we could estimate the outcome model by adjusting linearly for confounding; in addition, it is common (\citealp{wooldridge_introductory_2012}, Chapter 9) to add a transformed (e.g., squared) variable to allow for some nonlinearities if theory suggests so (e.g., diminishing growth). This process is equivalent to the ``parametric" assumption that the functional form is known and correctly stated by the researcher \citep{imbens_nonparametric_2004}. Also, researchers typically select the variables which enter the estimation based on theory, domain knowledge, or intuition, which relies on assumptions as well. In some cases, all of these assumptions might be correct, but they are usually untestable, rather subjective or even arbitrary and different assumptions might lead to very different causal estimates.

When facing these challenges --- nonlinearities and variable selection problems --- it is natural to think about using (supervised) ML. One of the well-known strengths of many ML methods is their ability to deal with high-dimensional data and to fit complex functional forms \citep[e.g.,][]{mullainathan_machine_2017, athey_impact_2019, hastie_elements_2009}.
However, a ``naive" application of ML methods for these causal questions leads to biased estimates, since the goal of traditional ML is prediction, which is fundamentally different from causal parameter estimation \citep{mullainathan_machine_2017}. The reason is that ML methods use regularization, which keeps predictive models from overfitting, but introduces a bias into parameter estimates \citep{chernozhukov_doubledebiased_2018}.

At this point, the DML method comes in and suggests an estimation procedure which can use highly flexible, regularized ML methods while still providing an unbiased estimator for the causal effect. As a consequence, we can relax assumptions about functional forms and variable selection. Algorithm \ref{alg:dml_plr} depicts a version of the DML algorithm for the partially linear regression model \citep{chernozhukov_doubledebiased_2018}. In other settings (e.g., binary treatment in the interactive model, instrumental variables, etc.), the algorithms and final estimators can be slightly different (see Section \ref{sec:furth_settings}). 
In the first step of the algorithm, we split the full dataset randomly into $K$ folds. Secondly, we hold out one of these folds, and use the remaining $K-1$ folds to train two ML models: The first model predicts the treatment $W$ from the potential confounders $\boldsymbol{X_c}$; the second model uses the same variables to predict the outcome $Y$. In step three, we use these trained models to make predictions for treatment and outcome, respectively, on the data not used for training. Then, we subtract these predictions from the true values to obtain the residuals for treatment and outcome ($\hat{V}_W$ and $\hat{V}_Y$). Next, we linearly regress the residual of the outcome ($\hat{V}_Y$) on the residual of the treatment ($\hat{V}_W$) and obtain the coefficient for $\hat{V}_W$. Finally, we repeat steps 2-5 for each of the $K$ folds, resulting in $K$ different coefficients, which we finally average to obtain the final causal estimate.\footnote{\citet{chernozhukov_doubledebiased_2018} name this algorithm ``DML1". There is an alternative algorithm ``DML2", which supposedly could perform better in small samples. Instead of computing a coefficient for each fold, DML2 collects the residuals across all folds and estimates the final effect on the full, residualized dataset. In our simulations, we could not confirm a superior performance of DML2, therefore we continue with the more intuitive DML1.} 
Because the random splitting in the first step can potentially influence the estimation results, there is an additional step for more robustness in smaller samples: \citet{chernozhukov_doubledebiased_2018} recommend repeating the full algorithm multiple (e.g., 100) times with different partitions, and finally reporting the median estimate across all splits.

\begin{algorithm}[ht]
  \caption{DML algorithm for the partially linear regression model}
  \label{alg:dml_plr}
  \setlist{nolistsep}
  \begin{enumerate}[leftmargin=*]%, noitemsep]
  \itemsep0.3em 
  \vspace{.3em}
    \item Split the data into $K$ folds
    \item Train two machine learning models on $K-1$ folds:
    \begin{enumerate}[label=\alph*)]
      \item Outcome: $W$, features: $\boldsymbol{X_c}$
      \item Outcome: $Y$, features: $\boldsymbol{X_c}$
    \end{enumerate}
    \item Use the models to make predictions ($\hat{W}$ and $\hat{Y}$) on the held-out fold
    \item Compute residuals as $\hat{V}_W = W - \hat{W}$ and $\hat{V}_Y = Y - \hat{Y}$
    \item Use a linear regression to estimate coefficient from residuals:  
    Regress $\hat{V}_Y$ on $\hat{V}_W$, obtain the coefficient on $\hat{V}_W$
    \item Repeat for all folds, average resulting coefficients to obtain the final causal estimate
    \item For more robustness w.r.t. the random partitioning in finite samples: \\ %$\rightarrow$ 
    Repeat the algorithm $S$ (e.g., 100) times for different splits, then report the median estimate
    \vspace{.3em}
  \end{enumerate}
\end{algorithm}

DML relies on two steps to obtain causal effect estimates. First, it estimates a model for both the outcome and the treatment. As a consequence, the final estimator is robust to minor mistakes in the estimation of either of these models, whereas traditionally, one would have to assume that the outcome model was correctly specified. The authors call this technique ``orthogonalization", and it is closely related to the ``double robustness" property of other estimators (e.g., \citealp{hernan_causal_2020}, Chapter 13). Doubly robust estimators rely on estimating two models from the covariates: a treatment model and an outcome model. Combining the two models in the final estimator leads to the following robustness property: The estimator is consistent, as long as one of the two models is correctly specified.  That is, even if one model is misspecified (wrong functional forms, missing variables), but the other one is correctly specified, the estimates will still be valid  (\citealp{hernan_causal_2020}, Chapter 13). The term orthogonalization then comes from the fact that in DML, we use the residuals for the final estimation. These residuals have been constructed to be orthogonal to the confounders (i.e., cannot be explained by the confounders). That is, by subtracting the predicted outcome and treatment values from their true values, we approximately remove the confounding influence in both variables. This estimator is then doubly robust in that the errors of both models enter the overall estimation error as a product, such that if one error goes towards zero, the overall error does so as well. Thus, even though the regularization bias of using ML directly in the outcome model is too large to ignore, we can get closer to unbiased estimates if we also consider the treatment model and combine both models in a doubly robust way, as DML does. Please see Section 2 of \citet{chernozhukov_doubledebiased_2018} for a formal derivation of this approach.

The second step is an efficient kind of sample splitting, which the authors call ``cross-fitting". The goal of cross-fitting is to remove a bias caused by overfitting, which can occur if we use the same data to fit the ML models and to estimate the causal effect. Thus, in cross-fitting, we split the data into (at least) two samples: one for training the ML models, the other one for predicting, computing the residuals, and estimating the causal effect. This sample-splitting generally leads to a loss of efficiency, because we only use a part of the data in each step. However, cross-fitting regains efficiency by swapping the samples and repeating the procedure, such that we use each sample once for fitting the model and once for estimating the causal effect. This leads to multiple estimates (one for each sample) which we finally average.

Using DML may allow researchers to relax variable selection and functional form assumptions. Traditionally, a researcher would assume a prespecified set of confounders with prespecified (e.g., linear) functional forms. If these variable selection and functional form assumptions are violated, traditional methods only remove a part of the confounding variation, while the remaining confounding variation will still bias the estimated effect. However, if we employ ML algorithms to learn the important variables and the appropriate functional forms, we can rely on a weaker assumption: We do not necessarily need to know the correct variables and functional forms beforehand, we only need to assume that the ML algorithms perform reasonably well at the two prediction tasks. By virtue of the double robustness property, a ``reasonably good" performance does not mean that the ML methods have to recover the true model perfectly; the method is insensitive to small mistakes in those models.\footnote{Technically, this means that we have to assume that the ML algorithms achieve $n^{1/4}$-consistency. Traditionally, if we assume a parametric model and it is correctly specified, we achieve $\sqrt{n}$-consistency. Loosely speaking, this means that, as the sample size $n$ grows, the error between the true and the estimated value shrinks to zero by the factor $1/\sqrt{n}$, which is faster than the factor $n^{-1/4}$. Since the DML method makes use of the double robustness property, the errors of the treatment and outcome model multiply. Thus, in order to achieve $\sqrt{n}$-consistency without having to make the parametric assumptions, we must assume that the ML methods achieve $n^{1/4}$-consistency in both models ($n^{1/4} * n^{1/4} = n^{1/2} = \sqrt{n}$) \citep{chernozhukov_doubledebiased_2018, belloni_program_2017}.}
This assumption is not directly testable, but it is well aligned with the objective of flexible ML methods and thus - in most settings - weaker than the traditional assumptions, if we do not have very strong theory for a specific parametric model.   

We emphasize that the DML algorithm alone does not by itself guarantee a causal interpretation of the estimates. There are important assumptions and decisions researchers need to make before applying the method or interpreting its results. Most importantly, the researcher still must assume a causal structure that implies unconfoundedness or another form of identification (e.g., instrumental variables). Typically, this assumption can be violated in two ways. 
First, there could be remaining unobserved or unmeasured variables that influence both treatment and outcome (Figure \ref{fig:dag_U}) \citep{imbens_recent_2009}. DML cannot reduce the bias induced by such unobserved confounders. DML can flexibly adjust for \textit{observed} confounders, but cannot handle confounders that are not observed. 
Second, researchers could unintentionally include a ``bad control" in the algorithm \citep[see also][]{hunermund_double_2023}. A bad control is any variable that we should not adjust for, because adjusting for it introduces a new bias \citep[e.g.,][]{angrist_mostly_2009, cinelli_crash_2022}. In the simplest case, a bad control can be a so-called ``collider", which is influenced by treatment and outcome instead of influencing them (Figure \ref{fig:dag_coll}). In practice, realistic cases of bad controls might be more complex \citep[see, e.g.,][]{pearl_book_2019, imbens_potential_2020}. 
As a consequence of these two potential pitfalls, before applying DML, researchers have to consider (1) whether there might be any important unobserved confounders and (2) whether every variable included is a good control. Both of these assumptions are not directly testable from data, so the researcher has to argue from theory and domain knowledge about their plausibility. 
 
\begin{figure}[ht]
\centering
\begin{subfigure}[t]{0.45\textwidth}
\centering
\begin{tikzpicture}
    \node (w) at (0,0) [label=left:$W$,point];
    \node (y) at (6,0) [label=right:$Y$,point];
    \node (xc) at (3,1) [label=above:$\boldsymbol{X_c}$,point];
    \node (u) at (3, -1) [fill=white, label=below:$U$,point];
    
    \path (w) edge (y);
    \path (xc) edge (y);
    \path (xc) edge (w);
    \path[dashed] (u) edge (y);
    \path[dashed] (u) edge (w);    %\path (xp) edge (y);
\end{tikzpicture}
\caption{The unobserved confounder $U$ biases the estimated effect.}
\label{fig:dag_U}
\end{subfigure}
\begin{subfigure}[t]{0.05\textwidth}
    \hspace{0.05\textwidth}
\end{subfigure}
\begin{subfigure}[t]{0.45\textwidth}
\centering
\begin{tikzpicture}
    \node (w) at (0,0) [label=left:$W$,point];
    \node (y) at (6,0) [label=right:$Y$,point];
    \node (xc) at (3,1) [label=above:$\boldsymbol{X_c}$,point];
    \node (xcoll) at (3, -1) [label=below:$\boldsymbol{X_{coll}}$,point];

    \path (w) edge (y);
    \path (xc) edge (y);
    \path (xc) edge (w);
    \path (y) edge (xcoll);
    \path (w) edge (xcoll);
    %\path (xp) edge (y);
\end{tikzpicture}
\caption{Adjusting for bad controls like collider $X_{coll}$ introduces a bias.}
\label{fig:dag_coll}
\end{subfigure}
\caption{Possible violations of unconfoundedness.}
\label{fig:dagviol}
\end{figure}

\subsection{Possible ML algorithms}

The underlying statistical theory of DML allows researchers to use of a variety of different predictive or ML methods for the treatment and outcome model \citep{chernozhukov_doubledebiased_2018}. The authors mention algorithms such as ``random forests, lasso, ridge, deep neural nets, boosted trees, and various hybrids and ensembles of these methods". 
In our implementations of DML, we will focus on the following predictive methods for the ML tasks:

\begin{itemize}
    \item linear regression, potentially with lasso regularization
    \item generalized additive models (GAMs)
    \item random forests
    \item boosted regression trees (XGBoost)
    \item neural networks
\end{itemize}

These methods differ in three dimensions: the flexibility to model various functional forms, the ability to perform variable selection, and the degree to which they need large sample sizes to achieve flexibility. 

First, tree-based methods like random forests and boosted trees, as well as neural networks with a nonlinear activation function, are inherently capable of fitting nonlinear functional forms, whereas linear regression and lasso are purely linear methods when applied to the original features. One can make the latter methods more flexible by transforming the features (e.g., with interactions and polynomials). However, these transformations lead to an increase in the numbers of parameters to be estimated, which is problematic for linear regression, as we discuss below. Also, researchers have to create the nonlinear transformations by hand with little theoretical guidance (which variables should interact to which degree? Which order of polynomials is sufficient? Which further functional forms might be useful?). GAMs fit multiple smooth functions of the features simultaneously, which enables them to flexibly model a variety of functional forms \citep{hastie_elements_2009}. However, this comes at a cost because the fitting of multiple functions also increases the dimensionality, which can be problematic if the number of raw variables is large relative to sample size. Furthermore, GAMs are limited to additive relationships and smooth functional forms, and they can only model interactions if we add them manually \citep{james_introduction_2021}. 

Second, lasso and tree-based methods like random forests and boosted trees can perform variable selection by default. That is, they will only make use of features that are predictive of the outcome, which enables them to work in high-dimensional settings with many variables or transformations. In contrast, linear regression, GAMs, and neural networks do not perform variable selection by default and thus fail in high-dimensional settings. There are approaches that enable a kind of variable selection in GAMs \citep[e.g.,][]{marra_practical_2011}, but by default, the method only works in low-dimensional settings (\citealp{hastie_elements_2009}, Chapter 9). Similarly, while neural networks do not select variables automatically, there are various other techniques to avoid overfitting, e.g., early stopping, weight decay, dropout, and L1-regularization (e.g., \citealp[]{hastie_elements_2009}, Chapter 11). 

Lastly, while random forests, boosted trees, and neural networks can very flexibly fit different functional forms, they might need relatively large sample sizes to do so (e.g., \citealp{hastie_elements_2009}, Chapter 10). 
On the other hand, linear methods do not become more flexible as the sample size increases. 

In this paper, we focus on the selected predictive methods on grounds of their respective ease of implementation in statistical software and programming languages. For some of the methods, we can tune parameters to improve predictive accuracy. In our implementations, we tune at most one parameter (e.g., in lasso, random forests, XGBoost, and neural networks). Our main goal is not necessarily to find the model that minimizes test error for each method, but to compare how different methods perform in the DML framework without substantial researcher input in the ML step. We specify parameter and tuning choices in the simulation section. In general, one can also consider further tuning of the methods, other algorithms, or ensembles of the mentioned methods, which might lead to further improvements in applications.

\section{Simulations} \label{sec:simulations}

In this section, we present and discuss the results from applying DML to a variety of simulated datasets. The goal is to assess the ability of the method to recover the true (simulated) causal effect in settings that differ with respect to the true data-generating process. First, we describe our implementation of DML, the different algorithms for the ML parts and the choice of parameters. Then, we begin the simulations with an arguably realistic baseline scenario. From there, we look for boundary conditions of the method by varying functional forms, numbers and kinds of variables, sample size, confounding strength, and causal structure. 
We generate 100 simulated datasets for each simulation setting and report how a variety of different methods perform across these 100 simulations.\footnote{We experimented with 50 to 1,000 simulation replications and found 100 to be fairly representative for the distribution of resulting coefficients, while still being relatively economical with computing resources. We show results for different numbers of replications in our baseline scenario in Appendix \ref{appx_nsim} (Figure \ref{fig:bynsim}).}

\subsection{Method implementations} \label{sim_implementation}

After having reviewed the theory behind DML in the previous section, we now present how we have implemented the different methods (Table \ref{tab:methods}) for our simulation study. 
We implement all methods in R \citep{r_core_team_r_2023}.\footnote{All our code is available on OSF \href{https://osf.io/eswfk/?view_only=0fb868c0d78e4550bf0a2ab5a71ce7b1}{here}.} 
As a benchmark for DML, we include a number of alternative methods. First, we estimate a ``Simple" OLS regression of the outcome on the treatment, completely ignoring any other variables. If confounding is present in the data, this method will lead to biased results, since it does not adjust for confounding at all. Secondly, we estimate another OLS regression, but this time adjust linearly for all observed covariates. If the true confounding relationships are linear and low-dimensional, this will adequately adjust for confounding \citep[see, e.g.,][]{imbens_causal_2015}. However, this approach cannot model any nonlinear relationships by default. Adding interactions and squared terms for each covariate manually could help account for some nonlinearities, but also increases the number of variables in the model. Lastly, we implement one additional benchmark method that ``naively" uses ML (as described in the introduction of \citet{chernozhukov_doubledebiased_2018}), by only modeling the outcome model $g_0(\boldsymbol{X_c})$, neither relying on orthogonalization nor cross-fitting. To do so, we first obtain an estimate of $\beta$ by linearly regressing $Y$ on $W$. Then, we get an estimator $\hat g_0$ by regressing $Y - \hat \beta W$ on $\boldsymbol{X_c}$ using an ML method. Lastly, we use this estimator to predict the original outcome $Y$, compute the outcome residual and regress the residual on the treatment $W$ to obtain an effect estimate. We implement this naive ML method with XGBoost as the ML algorithm, selecting the parameters as described below. The results of other ML methods in this naive framework are very similar, so we focus on the version with XGBoost as representative.

\begin{table}[ht]
\caption{Description of implemented methods}
\scriptsize
\centering
\begin{tabular}{@{}ll@{}}
\toprule
\textbf{Label} & \textbf{Description}                                       \\ \midrule
Simple OLS & Ordinary least squares estimation of a linear regression model without any covariates                       \\
OLS & OLS estimation of a linear regression model with all covariates included linearly    \\
XGBoost (naive) & XGBoost without orthogonalization and cross-fitting        \\
OLS (DML) & DML with OLS regression as predictive algorithm         \\
Lasso (DML) & DML with lasso as predictive algorithm                     \\
GAMs (DML) & DML with GAMs as predictive algorithm                      \\
Random forests (DML) & DML with random forests regression as predictive algorithm \\
XGBoost (DML) & DML with XGBoost as predictive algorithm                   \\ 
Neural nets (DML) & DML with neural networks as predictive algorithm \\
\bottomrule
\end{tabular}%
\label{tab:methods}
\end{table}

We implement all versions of DML as described in Algorithm \ref{alg:dml_plr} (DML1 of \citet{chernozhukov_doubledebiased_2018}). As defaults, we use $K = 5$ folds for the cross-fitting procedure and  repeat the full algorithm nine times for different sample splits ($S = 9$), from which we take the median estimate. The authors recommend $K = 5$; and $S = 9$ is a number of repetitions that leads to stable estimates in our simulations. We vary these parameters in later simulations.   
For the ML parts of the algorithm, we use different predictive methods: OLS regression, lasso, generalized additive models (GAMs), random forests, boosted trees (as implemented by XGBoost), and neural networks: 

\begin{enumerate}
    
\item We implement lasso with the \textit{glmnet} package by \citet{simon_regularization_2011}. We determine the regularization parameter $\lambda$ via 10-fold cross-validation and choose it to minimize the cross-validation mean squared error on the $K-1$ folds used for training.  

\item For GAMs, we use the \textit{mgcv} package to model each continuous covariate with a smooth nonlinear function, estimated with restricted maximum likelihood (``REML")  \citep{wood_generalized_2017}. 

\item We implement random forests with the \textit{randomForest} package \citep{liaw_classification_2002}, train using 200 trees and tune mtry (the number of variables randomly sampled from at each split) with respect to Out-of-Bag error. 

\item For boosted trees, we utilize the implementation by the \textit{xgboost} package \citep{chen_xgboost_2023}. For the learning rate (eta) and the maximum tree depth, we use default values (0.3 and 6, respectively). Our implementation uses early stopping if the validation set performance does not improve for 10 rounds. We determine the optimal maximum number of boosting iterations by 5-fold cross-validation from up to 200 rounds and use it subsequently in the final model. 

\item Finally, we implement neural networks using the computationally very efficient \textit{nnet} package \citep{venables_modern_2002}. We use four units in one hidden layer with a sigmoid activation function and tune the parameter for weight decay between 0 and 5. Tuning the number of units and fixing the decay did not lead to better results. Tuning both the number of units and the parameter for the weight decay further improved the results, but led to an unreasonable increase in computational cost on CPUs (more than 7 times longer in our baseline than tuning one parameter or using DML with tuned random forests). In practice, if the computational resources are available, one can allow for more hidden layers and units, a different activation function, regularization techniques, and tune multiple parameters to achieve a good predictive fit. For our purpose, the above approach seems to achieve a good trade-off between predictive accuracy and computational cost, which makes the results more comparable to the alternative methods. 

\end{enumerate}

\subsection{Baseline simulation} \label{sim_baseline}

We assume the causal structure from Figure \ref{fig:dag1} for most simulations settings. That is, we can estimate the causal effect by adequately adjusting for all observed confounders (``unconfoundedness"). In later settings, we examine the consequences of a misspecified causal structure, i.e., we mistakenly leave out important confounders or adjust for bad controls. 

In all following simulations, we draw exogenous variables from a multivariate normal distribution with a mean of zero and a randomly generated covariance matrix (e.g., $X_c \sim N(0, \Sigma)$, $\Sigma = A'A$, with $A \sim N(0, 0.5)$).
The true causal coefficient of interest is 1 ($\beta = 1$). We draw intercepts and noise terms from a standard normal distribution ($\alpha, \epsilon \sim N(0,1)$). In our baseline scenario, we construct treatment $W$ and outcome $Y$ based on Equations \ref{eq:sim_treat} and \ref{eq:sim_out}.
We use functional forms with some complexity that are could plausibly occur in nature: a linear form, higher-order polynomials, an interaction, and a step-function. 
\begin{gather}
W = \alpha_0 + \delta_1 X_1  + \delta_2 {X_2}^2  + \delta_3 X_1 X_2  + \delta_4 step(X_3)  + \delta_5 {X_4}^3  + \epsilon_0  \label{eq:sim_treat} \\
Y = \alpha_1 + \beta W + \gamma_1 X_1  + \gamma_2 {X_2}^2  + \gamma_3 X_1 X_2  + \gamma_4 step(X_3)  + \gamma_5 {X_4}^3  + \epsilon_1  \label{eq:sim_out}
\end{gather}
Thus, the variables $X_1$ to $X_4$ are confounders that one should adjust for. We set the confounding strength $\delta_j = \gamma_j = 0.1$. $step()$ is the implementation of a step function, where we draw the initial variable from a standard normal distribution, before assigning each value to one of four steps between -1 and 1, depending on the quartile the value belongs to.

\begin{figure}[ht]
    \centering
    \includegraphics[width=.8\textwidth]{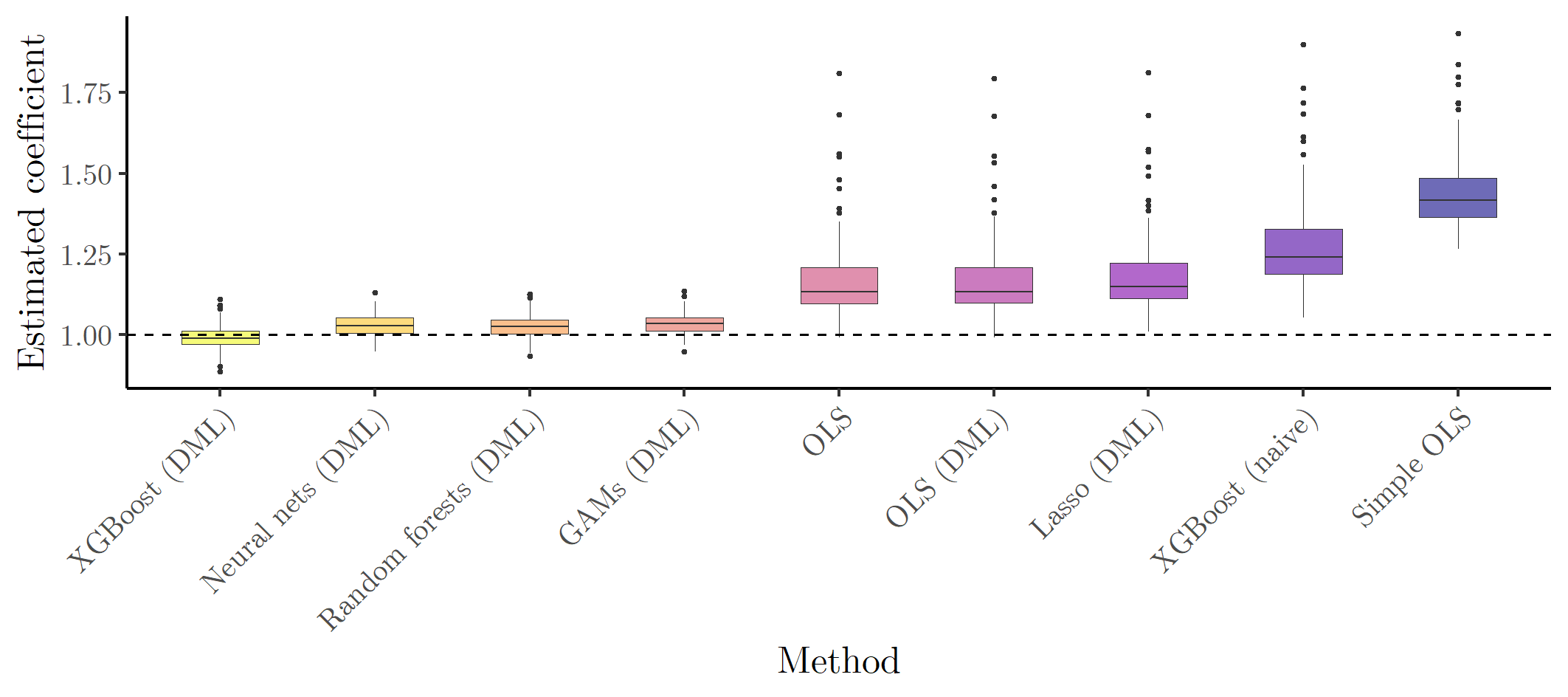}
    \caption{\label{fig:main}Results for our baseline simulation with sample size $n=1000$. The horizontal axis displays the different methods from Table \ref{tab:methods}. The vertical axis depicts the estimated coefficient. The dashed line marks the true causal effect ($\beta = 1$). The boxplots show the distribution of estimated coefficients across 100 simulated datasets for each method.}
\end{figure}

The results of our baseline simulation demonstrate the ability of DML with flexible ML methods to appropriately adjust for the various confounding influences (Figure \ref{fig:main}).
As expected, the naive simple OLS regression performs worst, due to its inability to adjust for confounding in any way. The naive ML method is able to adjust for some confounding, but still incurs significant bias due to regularization bias and overfitting. This is evidence for the notion that we should not use simple ML tools for causal inference. The traditional OLS regression, adjusting for all covariates linearly, outperforms the naive ML method, but is still biased. 

Within the DML framework, the linear methods perform very similarly compared to the standard OLS regression. By contrast, the inherently flexible ML methods are most successful at adjusting for the observed confounding. The median estimates of DML with  GAMs, random forests, neural networks, and  XGBoost are only off by about 3\%, 2\%, 2\%, and 1\%, respectively. These methods seem to be able to learn the functional form of the confounding from the data, without the need to include transformed covariates. 

\subsection{Extended simulations} \label{sim_further}

In the following, we extend our simulations to eleven further scenarios to assess DML's value and limitations (Table \ref{tab:sim_ext}).

\def\arraystretch{1.2}

\begin{table}[!ht]
\caption{Description of simulation scenarios}
\newcolumntype{a}{X}
\newcolumntype{k}{>{\hsize=.51\hsize}X}
\newcolumntype{h}{>{\hsize=1.49\hsize}X}
\scriptsize
\centering
\begin{tabularx}{\textwidth}{ckah}
\toprule
\textbf{Case} & \textbf{Category}              & \textbf{Data-generating process}                              & \textbf{Goal}                                                                    \\ \midrule

1    & Functional form       & Confounding influence can take on different functional forms: Linear, U-shaped, interactions, cubic, step function, or random                          & Examine performance of methods for various functional forms that differ in the degree of nonlinearity, smoothness, and additivity                     \\
2   & Confounding strength & Multiplying the confounding coefficients by larger integers                      & Assess impact of stronger confounding on estimates of methods  \\
3   & Confounding dimension & Varying numbers of confounders                       & Assess robustness of methods to larger numbers of important confounders   \\
4   & Sample size           & Varying the sample size                              & Examine data size requirements of methods                                  \\ 
5   &  Inclusion of further variables     & Including varying numbers of noise variables & Assess robustness of methods to the inclusion of irrelevant variables      \\ 
6   & & Including variables only related to outcome            & Learn whether inclusion of these variables is beneficial or harmful  \\ 
7   &     & Including variables only related to treatment            & Learn whether inclusion of these variables is beneficial or harmful  \\ 
8 &
  Violations of identification &
  One unobserved confounder in addition to observed confounders &
  Assess robustness of methods to unobserved confounding \\
9    &                       & Covariates are colliders instead of confounders & Examine consequences of misclassifying variables as good controls         \\ 
10    & Parameters in DML        & Varying the number of folds $K$ into which we split the sample & Examine impact of choosing $K$ and distributing observations between prediction and estimation  \\ 
11   &                       & Varying the number of algorithm repetitions $S$ & Examine additional robustness achieved by repeating DML        \\ \bottomrule
\end{tabularx}
\label{tab:sim_ext}
\end{table}

For the first five cases, we slightly deviate from the basic setup of the baseline's DGP. Instead of keeping all confounding coefficients fixed to 0.1, we randomly draw them from a standard normal distribution ($\delta_j, \gamma_j \sim N(0,1)$). We constrain $\delta_j$ and $\gamma_j$ to have the same sign to rule out that confounding effects of multiple confounders cancel out. These new coefficients make the adjustment more challenging, both because of their higher average magnitude and because of the randomness of their magnitude. Similarly to \citet{chiang_multiway_2022}, \citet{belloni_high-dimensional_2014}, and \citet{mcconnell_estimating_2019}, the confounding coefficients decrease with each additional confounder. We achieve this by multiplying each confounding coefficient with $1/j$, for $j = 1, ..., J$ confounding variables. Finally, for all of the following simulation settings, we do not present results from using DML with OLS regression as predictive method (``OLS (DML)"), since this method is virtually equivalent to directly estimating an OLS regression (``OLS"). 

\subsubsection{Case 1} \label{case1}
In Case 1, we assess how well the different methods can adjust for different functional forms of the confounding influences $g_0()$ and $m_0()$. For now, we will assume that $g_0()$ and $m_0()$ have the same functional form (although with potentially different coefficients). The sample size is still $n = 1000$ and there are five confounders with the specified functional form. Table \ref{table:ffs} shows the equations that generate the different functional forms. The first functional form of the confounding is linear. Secondly, the confounders enter the model by a squared term and thus have an U-shaped influence. This nonlinear functional form commonly occurs in nature in the form of diminishing growth, e.g., the relationship between the amount of fertilizer and crop yield. Third, if there is more than one confounder, they can interact with each other, e.g., in pairs of two as we implement here. For each confounder, we randomly draw the identity of the other confounder it will interact with. This is the first non-additive and non-smooth functional form. Fourth, the influence of the confounders follows a non-smooth step function. This is challenging for methods only modeling smooth functions, but favors tree-based methods. To implement the step function, we first compute the quartiles for each confounder. Then, we transform the confounder by assigning to each original value a value from [-3, -1, 1, 3], depending on the quartile to which the original value belongs. The result is a monotonically increasing step function consisting of three steps. This function can occur when decision makers base their decisions on heuristics: e.g., when setting prices, a product manager might consider thresholds rather than a fully linear model of competitor prices. Only if a competitor price exceeds a certain threshold, the manager adjusts the own-price accordingly. Fifth, the confounding influence follows a cubic functional form. Cubic functions can occur in economics, when marginal costs fall with increasing output but rise again at some point \citep[e.g.,][]{beach_use_1949}. Since the cubic term leads to a very strong confounding influence, we scale it down by a factor of 0.25 to make it more comparable to the other functional forms. Sixth, we draw the functional form randomly for each confounder. We select from each of the first five forms with equal probability. This case is most similar to the baseline scenario, though more challenging because of the randomness in the confounding coefficients.

\def\arraystretch{1}%  1 is the default, change whatever you need

\begin{table}[ht]
\caption{The different options for the functional form of the confounding}
%\newcolumntype{a}{>{\hsize=1.75\hsize}X}
%\newcolumntype{k}{>{\raggedright\hsize=.25\hsize}X}
\centering
\scriptsize
\begin{tabular}{@{}OlOl@{}}
\toprule
\textbf{Label} & \textbf{Equation} \\ \midrule 
Linear     &     \(\displaystyle g(X_c) = m(X_c) = X_c \) \\  
U-shaped     &      \(\displaystyle g(X_c) = m(X_c) = {X_c}^2 \)  \\ 
Interactions     &    \(\displaystyle g(X_c) = m(X_c) = \sum_{p = 1}^P X_{c_p} X_{c_j} \quad \textrm{with} \quad j \in_R \{\{1, ..., P\} \setminus p\} \) \\  
Step     &      \(\displaystyle g(X_c) = m(X_c) =\begin{cases}
          -3  & \text{if} \, \quad X_c < Q_1 \\
          -1 & \text{if} \, \quad Q_1 \leq X_c < Q_2 \\
          1 & \text{if} \,  \quad Q_2 \leq X_c < Q_3 \\
          3 & \text{if} \,  \quad Q_3 \leq X_c \leq X_{c_{max}} \\
     \end{cases} \)  \\ 
Cubic     &      \(\displaystyle g(X_c) = m(X_c) = 0.25 * {X_c}^3 \)  \\ 
Random per confounder    &      Select for each confounder a functional form from above with $p = .2$ each. ($g_0(X_{c_p}) = m_0(X_{c_p})$)  \\\bottomrule
\end{tabular}%
\label{table:ffs}
\end{table}

\begin{figure}[ht]
    \centering
    \includegraphics[width=\textwidth]{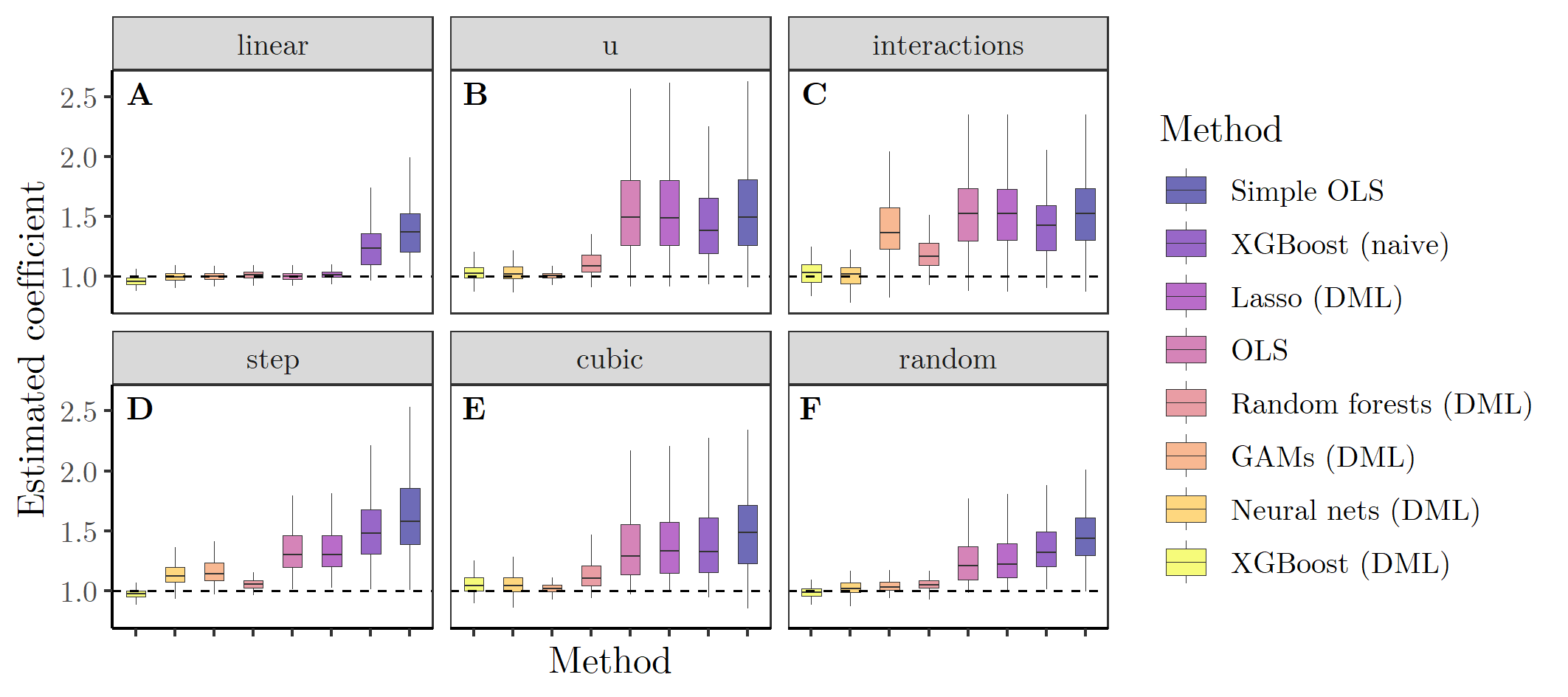}
    \caption{Results for Case 1 - distribution of estimated coefficients for each method across 100 simulations by functional form (outliers not displayed). The dashed line marks the true causal effect ($\beta = 1$). \textbf{A} Linear confounding. \textbf{B} U-shaped/squared confounding. \textbf{C} Pairwise interactions between confounders. \textbf{D} Confounding via step function. \textbf{E} Cubic confounding. \textbf{F} Confounding functional form drawn randomly for each confounder.}
    \label{fig:by_ff}
\end{figure}

The performance of the different methods varies significantly across the different functional forms (Figure \ref{fig:by_ff}). Since every simulation contains observed confounding, the method not adjusting for confounders at all (``Simple OLS") performs worst across scenarios, followed by the naive ML method. These methods' estimates vary widely across the 100 simulations, and their median displays a severe bias. 
When the confounding influence is linear (Figure \ref{fig:by_ff}A), all DML methods and the OLS regression perform well. 
DML with XGBoost is marginally biased in the direction opposite to the confounding, which potentially indicates minor overfitting. 

In case of a U-shaped confounding influence (Figure \ref{fig:by_ff}B), the linear methods are unable to adjust for confounding at all, thus providing results with bias similar to the simple OLS regression. The results demonstrate that although we can classify lasso as a ML method, it does not have any more flexibility than a classical OLS regression when it comes to fitting nonlinear functional forms from raw, untransformed variables. In this setting, DML with GAMs performs best, easily fitting the smooth quadratic function. DML with neural networks and DML with XGBoost have a slightly wider distribution of estimates, but are also close to unbiased at the median. DML with random forests is more biased, but still easily dominates the linear methods. 

Confounding through interactions (Figure \ref{fig:by_ff}C) is a non-smooth, non-additive functional form, hence purely linear methods are again unable to adjust for such confounding. GAMs in DML achieve more accurate results, but are still heavily biased. Both tree-based methods adjust better for interactions, although DML with XGBoost once more incurs the smaller bias. DML with neural networks most accurately estimates the effect and delivers unbiased estimates, slightly better than DML with XGBoost. 

Next, for the step function, linear methods are superior to not adjusting for confounding at all (Figure \ref{fig:by_ff}D). This is because we constructed the step function to increase monotonically. DML with GAMs gets even closer to the true coefficient, but cannot fit the steps well with smooth functions. Here, DML with neural networks is more accurate than DML with GAMs, but still incurs some bias and cannot compete with the tree-based methods. DML with random forests adjusts for confounding more successfully, and DML with XGBoost performs best, delivering estimates that are close to unbiased.

While the linear methods only adjust for small parts of the challenging, but smooth cubic confounding influence, DML with GAMs adjusts almost perfectly (Figure \ref{fig:by_ff}E). The tree-based methods and DML with and neural networks eliminate much confounding, but not as reliably as GAMs within DML.

If we draw the functional form randomly with equal probability for each confounder, the naive methods again perform worst, followed by the purely linear models (Figure \ref{fig:by_ff}F). DML with random forests, GAMs, neural networks, and XGBoost all deliver mostly accurate estimates, with DML with neural networks or XGBoost being closest to the true value most often. This result is very similar to our baseline simulation. 

Because the ``random" functional form is most similar to the baseline simulation, we use it in the simulations for Cases 2-5 and 10-11, where we vary other characteristics of the data-generating process. 

\subsubsection{Cases 2-4} \label{case2_4}
\begin{figure}[ht]
    \centering
    \includegraphics[width=\textwidth]{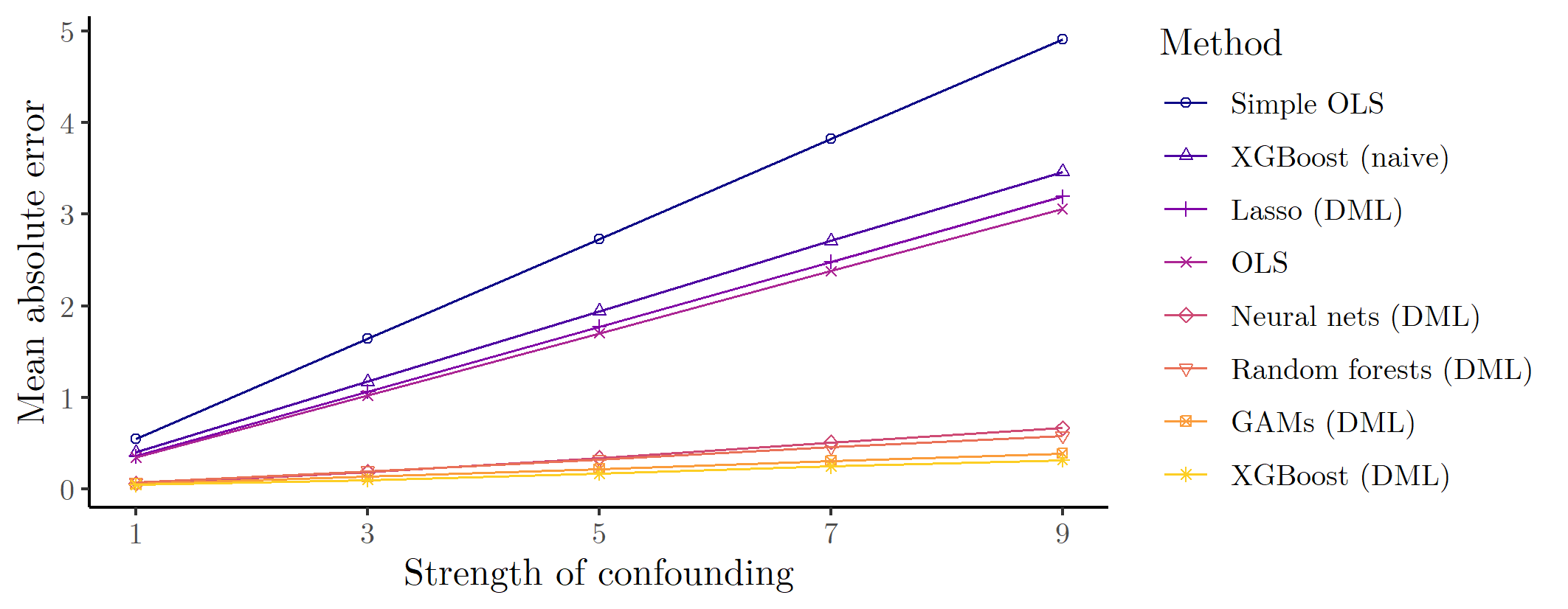}
    \caption{Results for Case 2 - mean absolute error in estimated coefficient across 100 simulations by varying confounding strength.}
    \label{fig:byconfStr}
\end{figure}

We visualize the impact of different characteristics in Cases 2-4 (and in Case 5) in terms of the mean absolute error (MAE) of the estimated coefficient. We chose MAE because it allows for a more accessible interpretation compared to the MSE. 
In Case 2, we vary the strength of the confounding influence by multiplying the confounding coefficient of the outcome ($\gamma_j$) with different positive integers. This increases the confounding bias and thus amplifies the importance of adequate adjustment.   
Increasing the strength of the confounding leads to a strongly increasing bias for methods that do not adjust well (Figure \ref{fig:byconfStr}). In fact, the MAE for all methods increases almost linearly, which is especially problematic for methods that are strongly biased in the first place. From this observation, we conclude that a larger confounding strength does not necessarily make the adjustment itself more difficult, but scales up the impact of the remaining confounding variation for all methods. The larger the confounding strength, the more important it is to adequately adjust for the confounding.

\begin{figure}[ht]
    \centering
    \includegraphics[width=\textwidth]{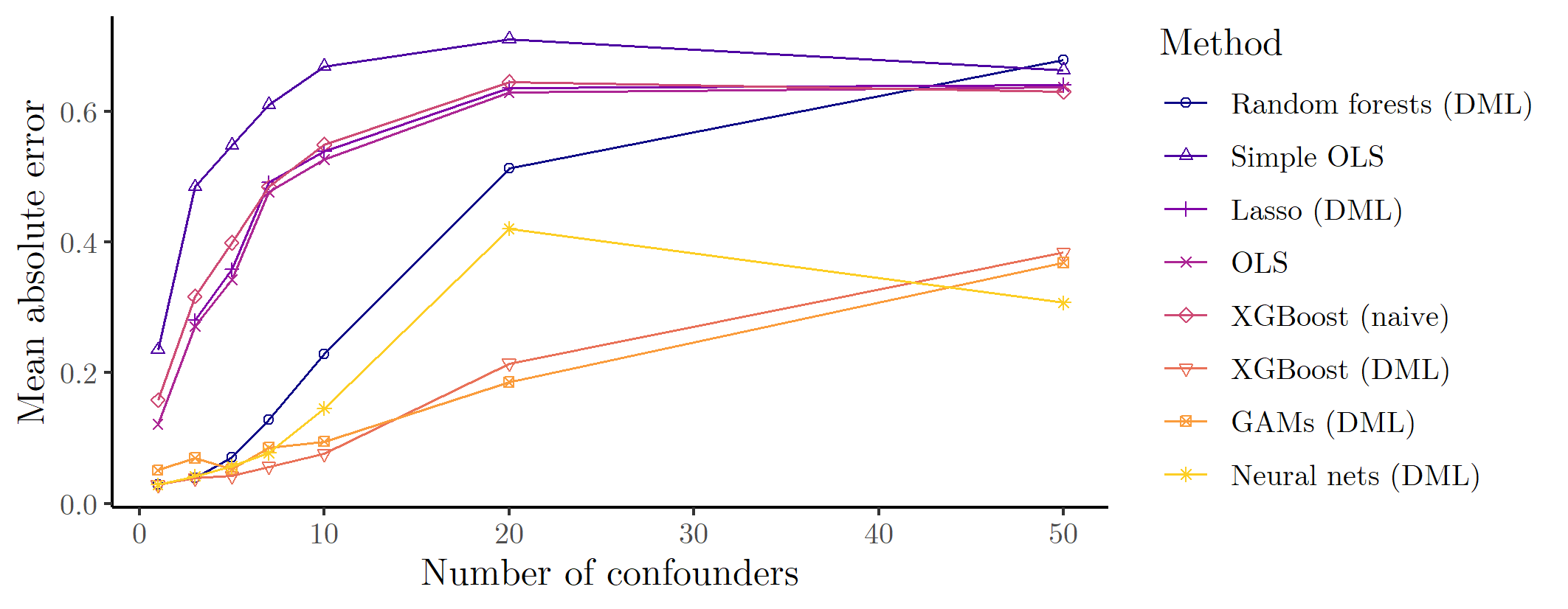}
    \caption{Results for Case 3 - mean absolute error in estimated coefficient across 100 simulations by varying numbers of confounders.}
    \label{fig:bynconf}
\end{figure}

In Case 3, we investigate how the number of confounding variables complicates the adjustment. All methods become less accurate as the number of confounders increases (Figure \ref{fig:bynconf}). The simple OLS regression remains worst, while the bias of the naive ML method and the linear methods becomes more similar as the number of confounders increase. The flexible DML methods are all reasonably accurate up to a number of five confounders. After that, their performance deteriorates, although at different paces. DML with random forests appears most sensitive to the inclusion of more confounders; at 20 confounders, it is slightly less biased than the linear methods; with 50 confounders, its estimates are essentially equally or more biased. DML with GAMs, XGBoost, or neural networks also incurs severe bias for very large numbers of confounders, but remain significantly more accurate than the alternative methods. Interestingly, DML with neural networks seems less affected by the inclusion of more than 20 confounders: for 50 confounders, the bias even somewhat decreases. 
These results demonstrate that the capability of DML to adjust for confounding in high-dimensional settings does not refer to the situation with many important raw confounders relative to sample size. Rather, DML can successfully adjust for complex functional forms, which it can model with a large number of parameters within flexible ML methods.

\begin{figure}[ht]
    \centering
    \includegraphics[width=\textwidth]{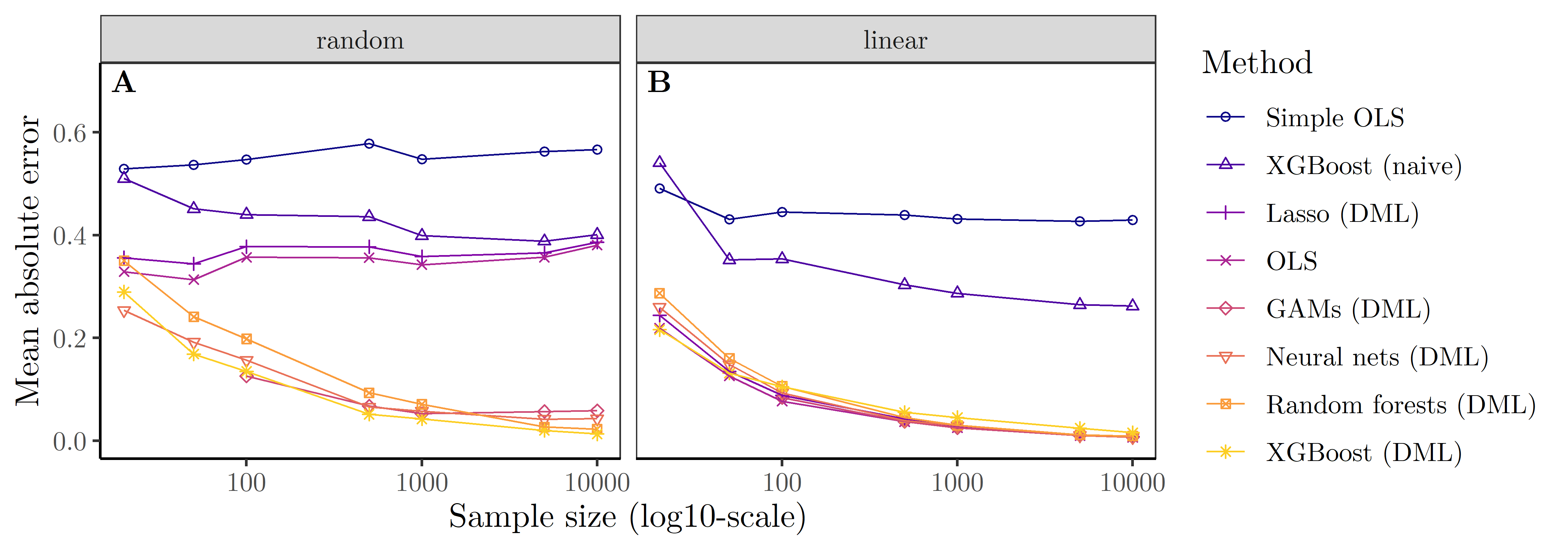}
    \caption{Results for Case 4 - mean absolute error in estimated coefficient across 100 simulations by varying  sample size. Sample size displayed in log10-scale. \textbf{A} Confounding with random functional forms. \textbf{B} Linear confounding.}
    \label{fig:bysample_size}
\end{figure}

While many ML algorithms are capable of flexibly learning functional relationships from data, most have certain functional forms they can model very easily, while they might need many more observations to learn others. For example, tree-based methods are good at modeling step-functions and interactions, but need more data points to model smooth or linear functions (see, e.g., \citealp{james_introduction_2021}, Chapter 8). Therefore, in Case 4, we vary the size of the simulated dataset. 
When the confounding influence has a random functional form (Figure \ref{fig:bysample_size}A), the simple OLS regression and the linear methods do not benefit from additional observations, since they are unable to learn nonlinear functional forms. By contrast, the naive ML method improves with larger sample size, but the improvement is much too slow to be substantial. For the flexible DML methods, the MAE decreases significantly with sample size, with the biggest improvements occurring between 20 and 500 observations. After this point, DML with GAMs does not improve further, whereas DML with random forests, XGBoost, or neural networks continues to learn and remove more of the confounding influence.  
In the case of linear confounding (Figure \ref{fig:bysample_size}B), we make similar observations. While the simple OLS regression does not benefit from additional observations, the naive ML method does, but again, the improvement is far too slow to realistically approach unbiased estimates. Since the confounding is linear, the linear methods and DML with GAMs perform best at any sample size. DML with random forests and neural networks have higher bias in very small samples, but they quickly learn the linear relationships as the sample size grows, ending up with estimates that are virtually indistinguishable from linear methods in large samples. 

\subsubsection{Cases 5-7} \label{case5_7}
So far, we have only considered scenarios where all covariates are confounders. However, there could also be noise variables $\boldsymbol{E}$, variables only influencing the outcome $\boldsymbol{X_p}$, and variables only influencing the treatment $\boldsymbol{X_z}$ (Figure \ref{fig:dagEXp}). In Cases 5-7, we explore the relevance of these variables for the performance of DML. First, in Case 5, we expect that methods capable of variable selection are less influenced by the inclusion of noise variables than others. Also, $\boldsymbol{X_p}$ and $\boldsymbol{X_z}$ should in asymptotic theory not impact the overall bias. However, in finite samples, adequately adjusting for $\boldsymbol{X_p}$ can increase precision (and decrease standard errors), while adequately adjusting for $\boldsymbol{X_z}$ can hurt precision (and increase standard errors) \citep{cinelli_crash_2022}. 

\begin{figure}[ht]
\centering
\begin{tikzpicture}
    \node (w) at (0,0) [label=left:$W$,point];
    \node (y) at (6,0) [label=right:$Y$,point];
    \node (xc) at (3,1) [label=above:$\boldsymbol{X_c}$,point];
    \node (e) at (-3,1) [label=above:$\boldsymbol{E}$,point];
    \node (xz) at (0,1) [label=above:$\boldsymbol{X_z}$,point];
    \node (xp) at (6,1) [label=above:$\boldsymbol{X_p}$,point];

    \path (w) edge (y);
    \path (xc) edge node[above, el] {$g_0(\boldsymbol{X_c})$} (y);
    \path (xc) edge node[above, el] {$m_0(\boldsymbol{X_c})$} (w);
    \path (xz) edge node[above, el, rotate = 90, xshift= -.65cm, yshift = -.2cm] {$l_0(\boldsymbol{X_z})$}(w);
    \path (xp) edge node[above, el, rotate = 90, xshift= .75cm, yshift = -.2cm] {$h_0(\boldsymbol{X_p})$}(y);
\end{tikzpicture}
\caption{DGP including noise variables $\boldsymbol{E}$, variables only influencing the treatment ($\boldsymbol{X_z}$), and variables only influencing the outcome ($\boldsymbol{X_p}$). The influence of $\boldsymbol{X_z}$ and $\boldsymbol{X_p}$ is potentially complex and nonlinear ($l_0(\boldsymbol{})$, $h_0(\boldsymbol{})$).}
\label{fig:dagEXp}
\end{figure}

\begin{figure}[ht]
    \centering
    \includegraphics[width=\textwidth]{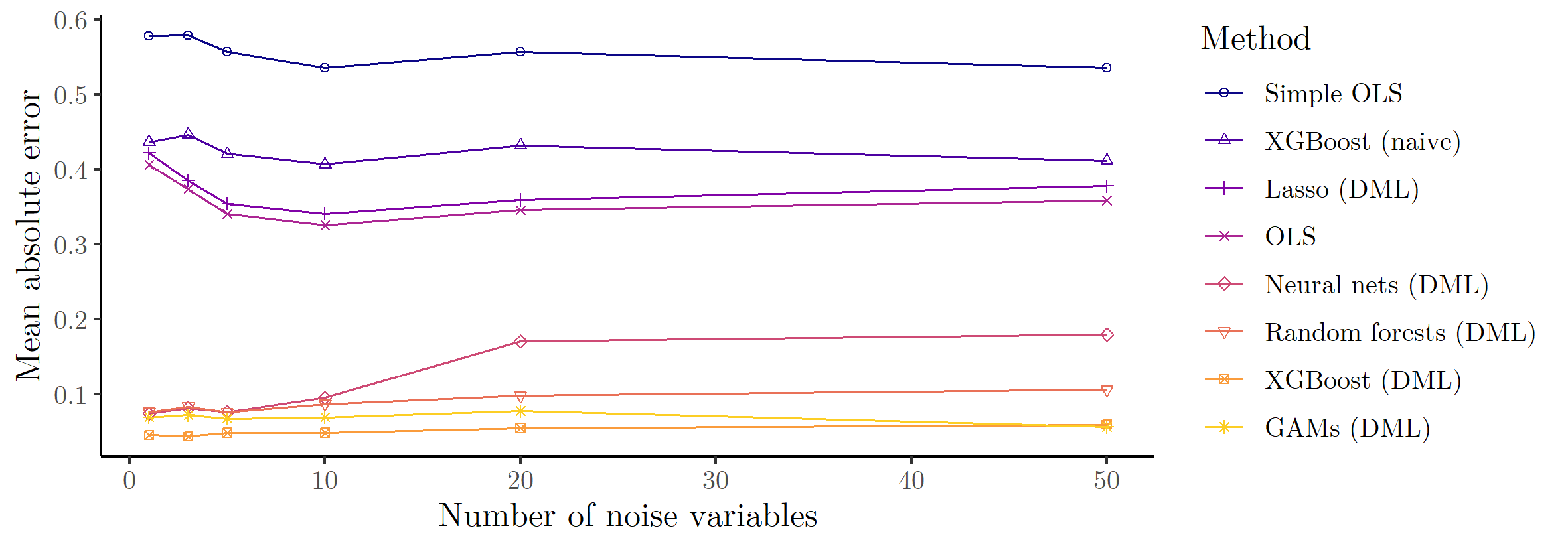}
    \caption{Results for Case 5 - mean absolute error in estimated coefficient across 100 simulations by varying number of noise variables.}
    \label{fig:bynoise}
\end{figure}

The inclusion of noise variables (Figure \ref{fig:bynoise}) affects none of the methods severely, except for DML with neural networks. The high flexibility of neural networks, combined with their inability for variable selection in our implementation, likely causes them to overfit the noise variables (which could potentially be avoided by further tuning and regularization). Even though some other methods also do not perform variable selection, they can still handle moderate numbers of noise variables. An OLS regression, for example, simply estimates near-zero coefficients for all noise variables. The main problem with high dimensionality arises when we try to make linear methods similarly flexible as, for example, ML methods like random forests, XGBoost, or neural networks. To do so, we could include various transformations of all variables (e.g., interactions and polynomials) which is problematic for an OLS regressions if the number of parameters approaches or exceeds the number of observations.

For the remaining cases, we return to the general setup of the baseline scenario, but change the nature of the variables we use. In Case 6, we introduce four additional variables ($\boldsymbol{X_p}$) which only influence the outcome. They take on the same functional forms as the confounders, but do not influence the treatment. We then estimate two models for each method: one adjusting for all variables ($\boldsymbol{X_c}$ and $\boldsymbol{X_p}$), the other only adjusting for the confounders ($\boldsymbol{X_c}$).

\begin{figure}[ht]
    \centering
    \includegraphics[width=\textwidth]{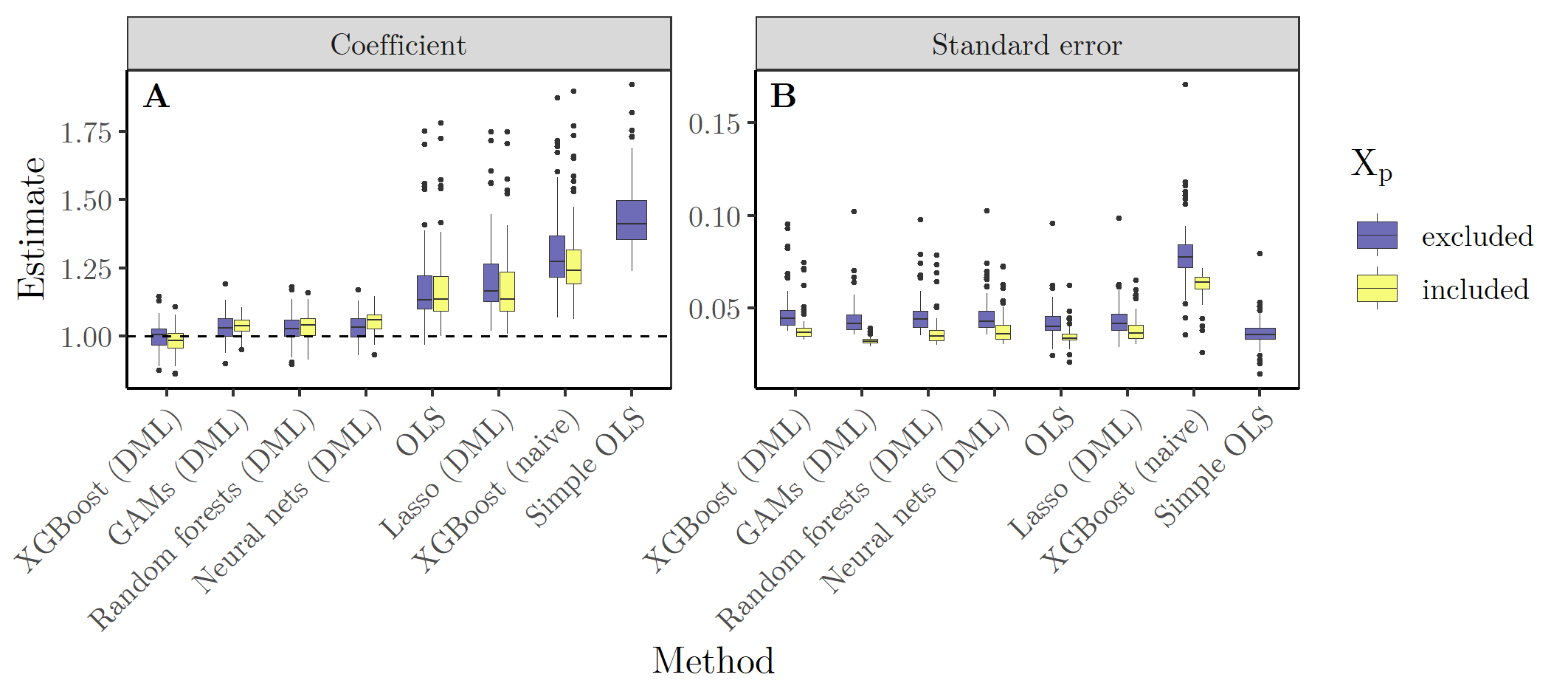}
    \caption{Results for Case 6 - distribution of \textbf{A} coefficient estimates and \textbf{B} standard error estimates across 100 simulations including (yellow) and excluding (blue) variables only related to the outcome. The dashed line marks the true causal effect ($\beta = 1$). Smaller standard errors are preferable.}
    \label{fig:xpplot}
\end{figure}

For most methods, adjusting for $\boldsymbol{X_p}$ in addition to $\boldsymbol{X_c}$ has little impact on the median coefficient estimate (Figure \ref{fig:xpplot}A). However, the boxplots tend to have a narrower distribution for the models that try to adjust for all variables. This in line with theory: By adjusting for $\boldsymbol{X_p}$, we reduce the variance in the outcome, which leads to more precise estimates \citep{cinelli_crash_2022}. The distribution of the standard errors  reinforces this finding: When adjusting for $\boldsymbol{X_p}$, the standard errors of all methods become significantly smaller (Figure \ref{fig:xpplot}B). From these results, we can derive the recommendation that researchers should adjust for all pre-treatment variables influencing the outcome, even if theory suggests that they are not related to the treatment. 

Different from the previous case, in Case 7, the additional covariates ($\boldsymbol{X_z}$) influence only the treatment instead of only the outcome. These variables again take on the same functional forms, and we once more estimate models adjusting and not adjusting for them.

\begin{figure}[ht]
    \centering
    \includegraphics[width=\textwidth]{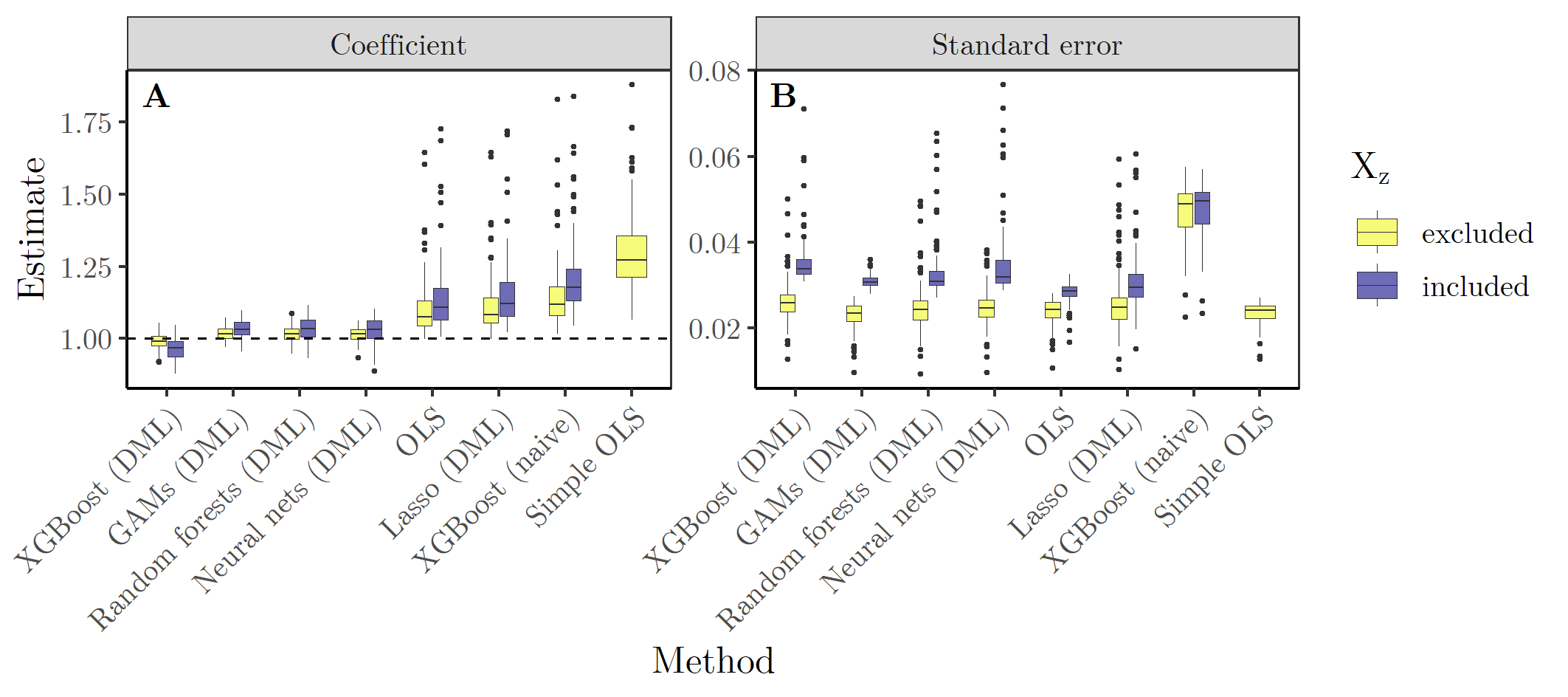}
    \caption{Results for Case 7 - distribution of \textbf{A} coefficient estimates and \textbf{B} standard error estimates across 100 simulations including (blue) and excluding (yellow) variables only related to the treatment. The dashed line marks the true causal effect ($\beta = 1$). Smaller standard errors are preferable.}
    \label{fig:xzplot}
\end{figure}

The results stand in contrast to the previous case: Now, the models that adjust for all variables, including $\boldsymbol{X_z}$, become less precise compared to the models that only adjust for confounders $\boldsymbol{X_c}$ (Figure \ref{fig:xzplot}A). In our simulations, this leads to both a more biased median and a wider distribution of estimates. Also, for most methods, including the variables related to the treatment leads to considerably larger standard errors (Figure \ref{fig:xzplot}B). Including these variables reduces the variance in the treatment, leaving less exogenous variation to estimate the effect \citep{cinelli_crash_2022}. Our results demonstrate that in finite samples, this can even lead to bias in the effect estimates. The bias should vanish asymptotically, but the loss of precision is sufficient reason to avoid including these variables in any models.

\subsubsection{Cases 8-9} \label{case8_9}
\begin{figure}[ht]
    \centering
    \includegraphics[width=.8\textwidth]{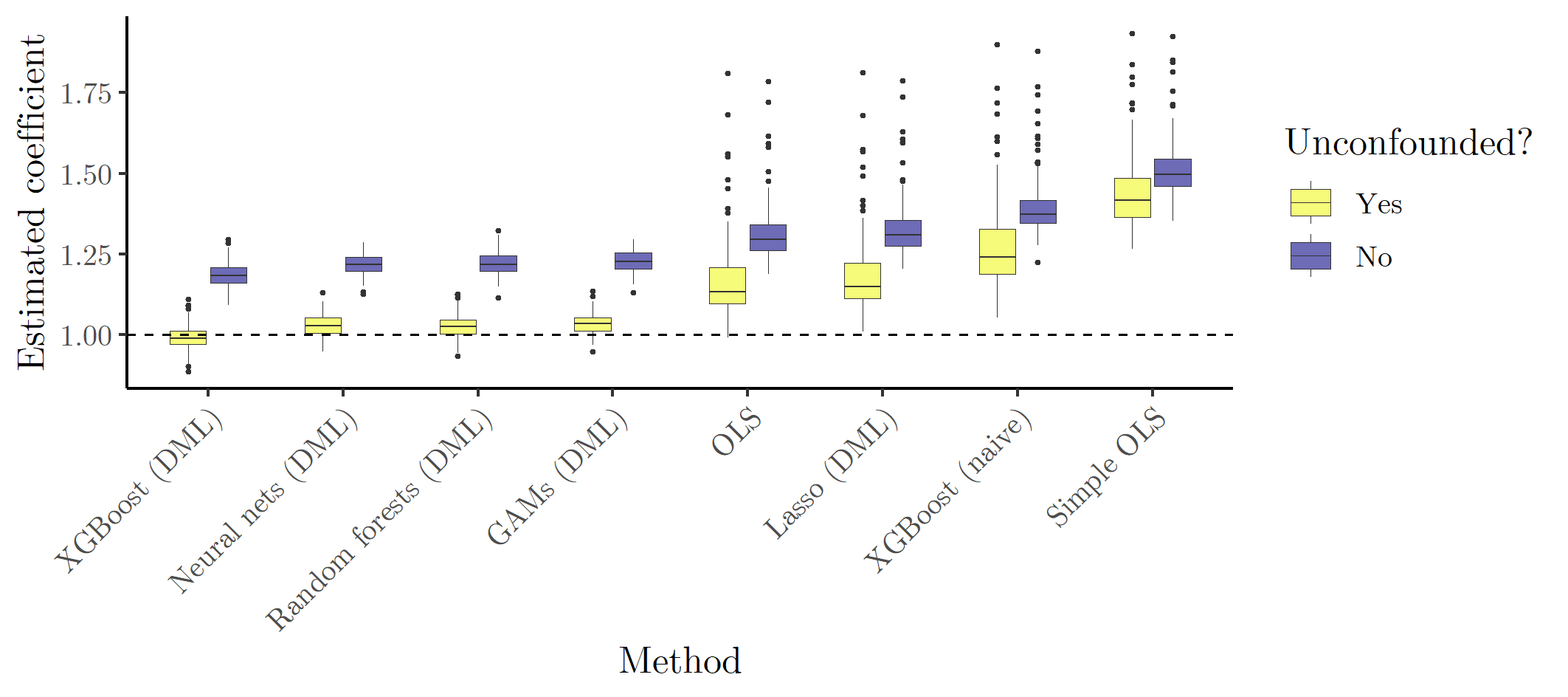}
    \caption{Results for Case 8 - distribution of estimated coefficients across 100 simulations without (yellow) or with (blue) unobserved confounding. The dashed line marks the true causal effect ($\beta = 1$).}
    \label{fig:unobs}
%\vspace{-2ex}
\end{figure}

As the final modifications to our DGP, we demonstrate in two different scenarios how violations to the assumed causal structure impact the estimates. So far in our simulations, assuming unconfoundedness and adjusting for all observed covariates was a valid identification strategy. Now, in Case 8, we include an additional confounder $U$ in the DGP from Equations \ref{eq:sim_treat} and \ref{eq:sim_out}, as visualized in Figure \ref{fig:dag_U}. Its coefficient is relatively large ($\delta_5 = 0.5$), but this confounder is not observed, hence we cannot adjust for it. 
The unobserved confounding affects all methods and makes them unable to recover the true causal effect (Figure \ref{fig:unobs}). Since $U$ is not observed, no method can adjust for it, irrespective of the method's flexibility. However, even though the estimates of all methods are biased, the bias is less pronounced for the flexible DML methods. This is not because they adjust for some part of the unobserved confounder, but because they still perform well at adjusting for the observed confounders.  So, even though unobserved confounding affects all methods, flexible DML might still \textit{relatively} outperform less flexible methods and lead to closer-to-truth estimates. This holds under the assumption that the unobserved confounding biases the estimate in the same direction as the observed confounding, otherwise the biases could cancel each other out. The plausible direction of the confounding can often be established from theory and domain knowledge for a specific application.

\begin{figure}[ht]
    \centering
    \includegraphics[width=.8\textwidth]{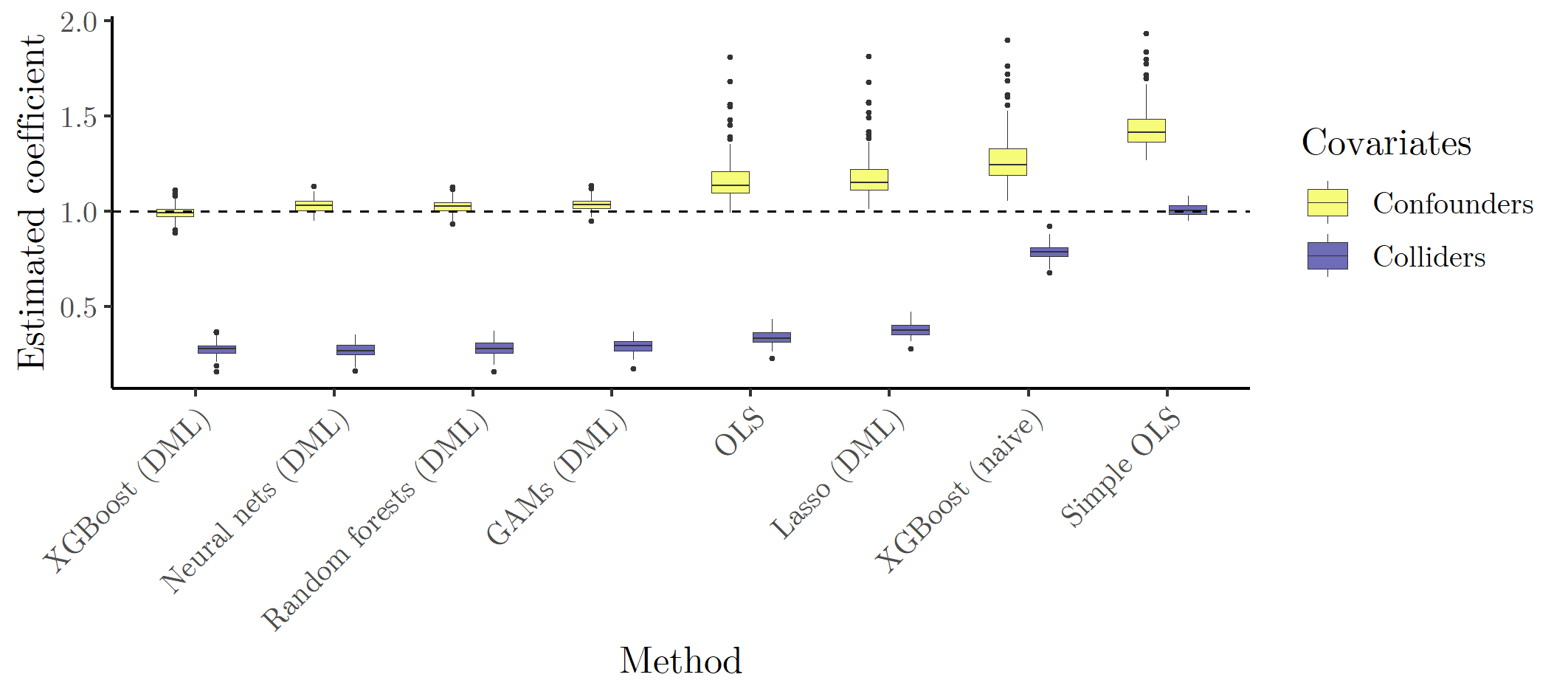}
    \caption{Results for Case 9 - distribution of estimated coefficients across 100 simulations when adjusting for confounders (yellow) or colliders (blue). The dashed line marks the true causal effect ($\beta = 1$).}
    \label{fig:coll}
\end{figure}

A second violation of unconfoundedness occurs when we adjust for a bad control, that is, a covariate that we should not adjust for. In the simplest case, a bad control is a collider variable as visualized in Figure \ref{fig:dag_coll}. To demonstrate the importance of the distinction between good and bad controls, we implement Case 9 such that all observed covariates are colliders instead of confounders. This means that there are five variables in $\boldsymbol{X_{coll}}$, caused by treatment and outcome with the same functional forms we used in the baseline. The simulation results confirm the theoretical statement that adjusting for bad controls leads to biased estimates (Figure \ref{fig:coll}). The only unbiased method is the simple OLS regression which does not adjust for any covariates. Interestingly, results of the baseline simulation are basically inverted if we adjust for colliders instead of confounders: The ``better" we adjust with flexible DML methods, the \textit{more} (downward) biased our results are. This emphasizes the importance of correctly classifying the covariates before entering them into any estimation algorithm. Since DML has no way of knowing whether a variable is a good or bad control, researchers have to make this decision based on theory and domain knowledge. These results caution against blindly throwing all available variables into DML or any other algorithm for causal effect estimation. 

\subsubsection{Cases 10-11} \label{case10_11}
\begin{figure}[ht]
    \centering
    \includegraphics[width=\textwidth]{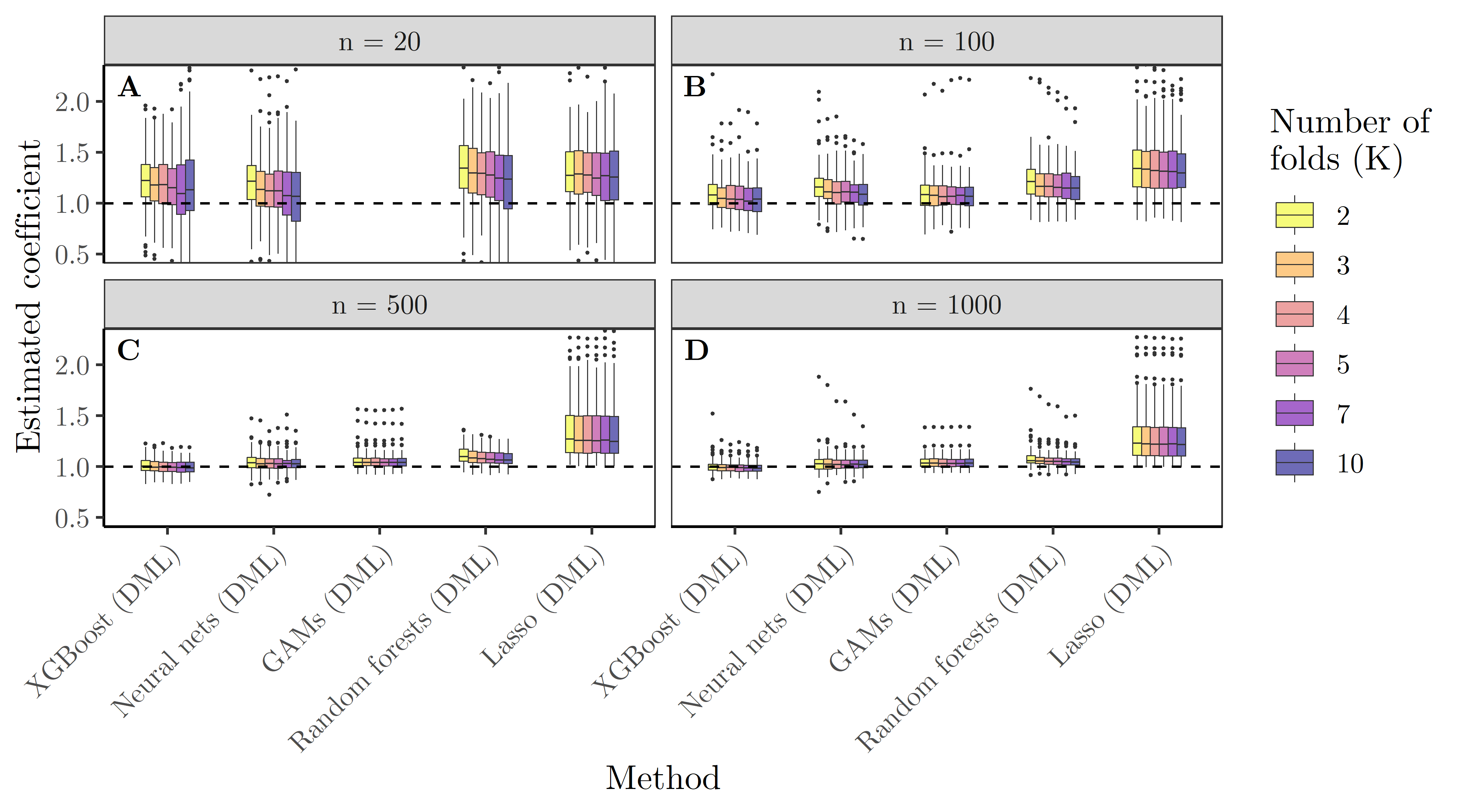}
    \caption{Results for Case 10 - varying the number of folds $K$ in DML for different sample sizes; distribution of estimated coefficients across 100 simulations. The dashed line marks the true causal effect ($\beta = 1$).
    \textbf{A} 20 observations. The number of parameters in flexible GAMs becomes too large for this sample size.  \textbf{B} 100 observations. \textbf{C} 500 observations. \textbf{D} 1000 observations.}
    \label{fig:byK}
\end{figure}

In the final Cases 10 and 11, we return to the baseline with the random functional form, but now we vary parameters of the DML algorithm. First, in Case 10, we vary the number of folds $K$ into which we split the dataset. \citet{chernozhukov_doubledebiased_2018} note that larger values of $K$ lead to larger samples for the ML step, which can help to increase predictive accuracy. At the same time, the sample for estimating the effect becomes smaller, but this step might depend less on the sample size. The authors mention values of 4 or 5 for $K$ as superior to $K=2$ in their tests, but we showed that 2 and 10 are the most common values for $K$ in applications (Figure \ref{fig:appl_lit}E; although $K=10$ is almost exclusively used in combination with lasso). 
For extremely small samples ($n = 20$), we observe that the median estimate of the flexible DML methods becomes more accurate as $K$ gets larger, but also that the variation in the estimates increases (Figure \ref{fig:byK}A). This is because the ML methods benefit from getting more observations for training; but for the extreme case of $K=10$, there are only two observations left for estimating the effect. For moderate sample sizes (Figure \ref{fig:byK}B and \ref{fig:byK}C; $n = 100$ and $n = 500$), the downsides of larger numbers of folds disappear and larger $K$s lead to both more accurate median estimates and smaller or at least similar variation in estimates. The same holds for larger samples (Figure \ref{fig:byK}D), but there the choice of $K$ is less influential, since even small $K$s provide sufficient observations to the ML algorithms. Similarly to before, we again observe here that only the ML methods that benefit from larger sample sizes are significantly affected by the choice of $K$. 

\begin{figure}[ht]
    \centering
    \includegraphics[width=\textwidth]{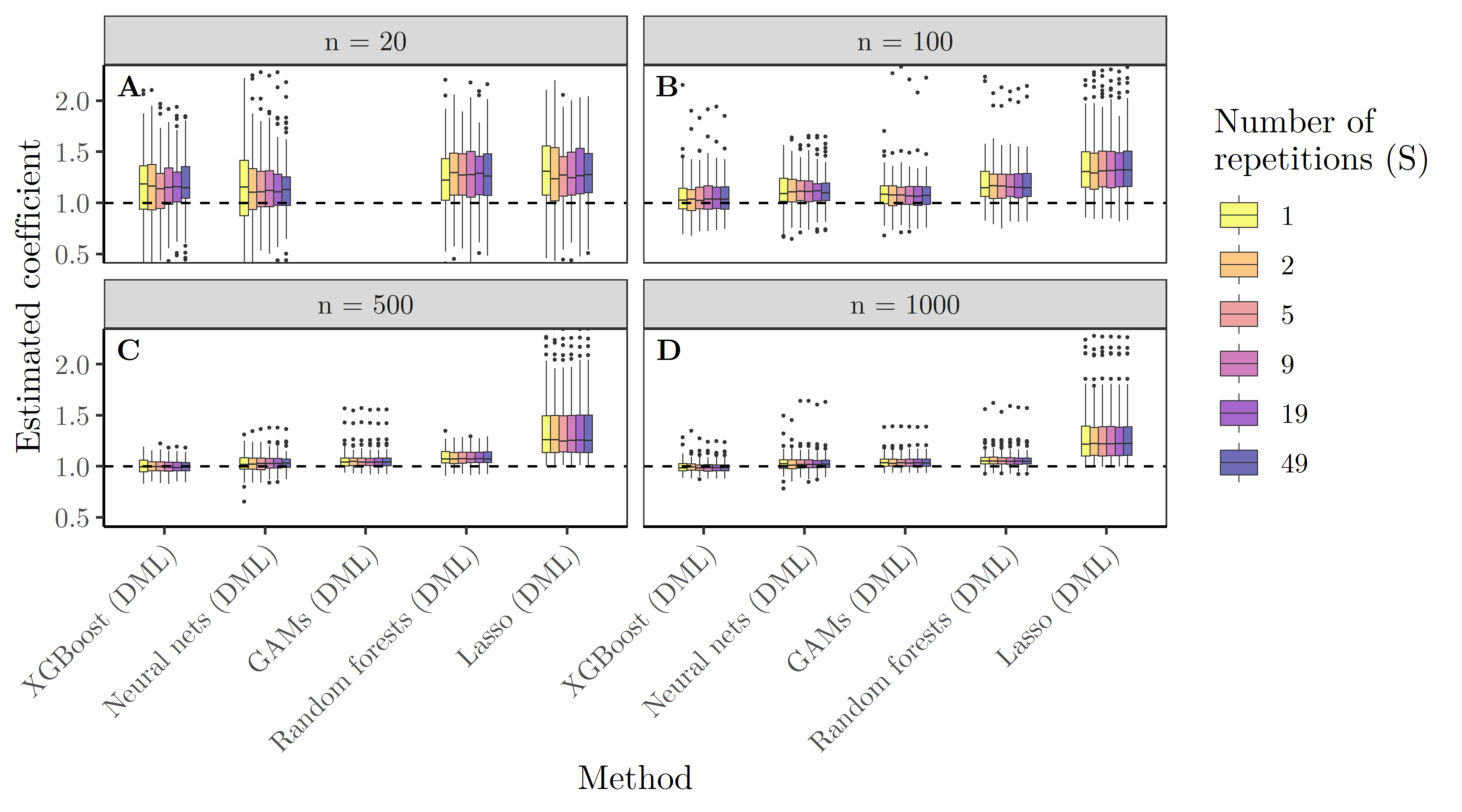}
    \caption{Results for Case 11 - varying the number of algorithm repetitions $S$ in DML for different sample sizes; distribution of estimated coefficients across 100 simulations. The dashed line marks the true causal effect ($\beta = 1$). \textbf{A} 20 observations. The number of parameters in flexible GAMs becomes too large for this sample size.  \textbf{B} 100 observations. \textbf{C} 500 observations. \textbf{D} 1000 observations.}
    \label{fig:bysplits}
\end{figure}

Secondly, we vary how often we repeat DML to make the estimates more robust to the randomness in the sample splitting. We repeat the algorithm $S$ times and report the median estimate from these $S$ estimates, which is more robust than the mean \citep{chernozhukov_doubledebiased_2018}. 
%shows the results for different numbers of sample splits on simulated datasets of different sizes. 
Our results show no obvious pattern in the accuracy of the median estimate: it remains relatively similar for larger numbers of repetitions (Figure \ref{fig:bysplits}). However, the variation of the estimates decreases moderately as the number of repetitions increases. This is most pronounced for very small samples, especially for DML with neural networks (Figure \ref{fig:bysplits}A). In larger samples, the benefit of many repetitions is negligible (Figure \ref{fig:bysplits}C-D). 
We will return to evaluating the benefits of multiple repetitions in the context of our application, where we observe a substantially larger impact of the choice of $S$. 

\subsection{Choosing between different ML algorithms} \label{sim_predictiveness}

While simulation results from the previous sections have given us an understanding of the capabilities of (versions of) DML in different scenarios, the choice between the different ML algorithms within DML is still not obvious, i.e., it is not obvious whether we should implement DML with, e.g., random forests, XGBoost, or neural networks. In the following analysis we investigate whether the accuracy of the different ML methods in predicting treatment and outcome, respectively, is indicative of the accuracy with which they can recover the causal effect. The underlying assumption is that a method which models the relationships between the confounders and treatment/outcome well will also be able to eliminate more of the confounding influence and thus deliver a less biased effect estimate compared to an estimator that is less accurate in predicting treatment and outcome. The cross-fitting procedure allows us to directly extract the predictive accuracy by averaging the errors across the respective folds that we held out from training. We use our baseline simulation to calculate the mean squared errors (MSE) for the prediction of treatment and outcome and the respective bias in the causal estimate.

\begin{figure}[ht]
    \centering
    \includegraphics[width=.6\textwidth]{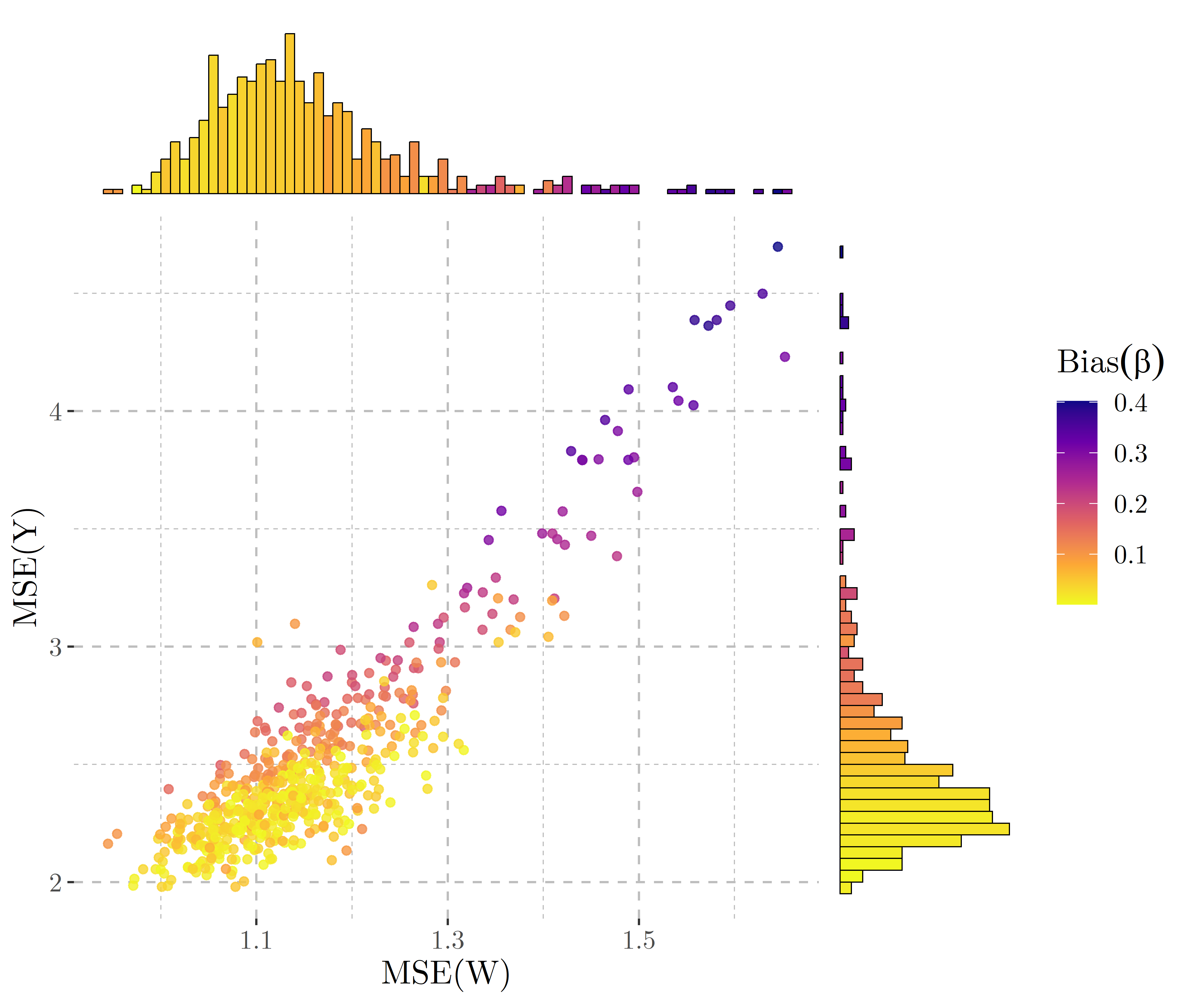}
    \caption{The relationship between the predictive accuracy of the first stage and the bias in the effect estimation.
    The horizontal axis shows the test MSE of the treatment prediction, the vertical axis the test MSE of the outcome prediction. The color scheme indicates the degree of bias, i.e., yellow represents a very small bias in the estimated causal effect, blue is relatively large bias. The histograms in the margins display the frequency distribution across the respective MSEs. We excluded the 2\% largest outliers in terms of MSE.
    }
    \label{fig:predictiveness}
\end{figure}

The results indicate a clear pattern: low values for the MSE(W) and the MSE(Y) are associated with yellow colors, i.e., small bias (Figure \ref{fig:predictiveness}). In other words, when the prediction of Y and W is good (small MSE), the bias in the causal estimate is small. Consequently, the scatterplot shows that the points in the bottom left corner have mostly small bias since both predictions are accurate, points in the upper right corner have mostly high bias because of inaccurate predictions.
Hence, under the assumption that we observe and adjust for all relevant confounders, we can use the predictive accuracy of the first stage as a guide for which estimate to trust more. If DML with two different ML methods gives very different results in terms of the estimated treatment effect, it seems reasonable to rely more on the one with smaller MSEs in the treatment and outcome predictions. Importantly, this comparison is only reasonable when the unconfoundedness assumption holds and when the models that we compare do not include any bad controls.

\section{Application to real-world data} \label{sec:application}

In this section, we use DML for an application that estimates the effect of one determinant of house prices. Building on \citeauthor{rosen_hedonic_1974}'s \citeyearpar{rosen_hedonic_1974} theory of hedonic prices, a variety of studies have applied hedonic price regressions to estimate the effects of particular attributes of goods or services \citep[e.g.,][]{agrawal_country_1999, nelson_residential_1978}. We replicate and extend the central regression model of \citet{harrison_hedonic_1978}, who use housing market data to measure the effect of air pollution on housing prices. As part of their analysis, they employ a rich cross-sectional dataset from the Boston area  (506 observations, 14 variables). They state that potential buyers know about the damages of air pollution and consequently are willing to pay more for a house in an area with low air pollution levels compared to an identical house in an area with high air pollution levels, and this higher willingness-to-pay would be reflected in a negative estimate for the effect of air pollution. 
\citet{harrison_hedonic_1978} argue that identification of this effect is possible with variation in air pollution over space after adjusting for a large number of observed neighborhood variables. 

\begin{figure}[ht]
\begin{minipage}[top]{\textwidth}
%\justifying
    \begin{center}
      \begin{tikzpicture}
      \small
          \node (w) at (0,0) [label=below:air pollution,point];
          \node (y) at (8,0) [label=below:house price,point];
        %   \node[align = center] (xc) at (3,3) {industry proportion \\ distance employment \\  zoned large lots \\ highways \\  crime rate \\ river};
          \node[align = center, fill = white] (xctext) at (4,1.7) {\% lower status \\river   \\ highways \\ crime rate \\   \% industry \\ zoned large lots  \\ distance employment };
          \node (xc) at (4, 2.5) {};
          \node[align = center] (xp) at (8,2.15) { pupil-teacher ratio \\ property-tax rate \\  \% blacks \\   rooms \\ age};
    
          \begin{pgfonlayer}{bg}  
              \path (w) edge (y);
              \path (xc) edge (y);
              \path (xc) edge (w);   
              \path (xp) edge (y);
          \end{pgfonlayer}
      \end{tikzpicture}
    \end{center}
\end{minipage}
\caption{The causal structure \citet{harrison_hedonic_1978} argue for when estimating the effect of air pollution on house prices. Some variables influence both treatment and outcome, others only affect outcome. The authors include all variables in a regression model of the outcome. We include the same variables in DML to predict both treatment and outcome and finally estimate the effect.}
\label{fig:dag_appl}
\end{figure}

\begin{figure}[ht]
    \centering
    \includegraphics[width=\textwidth]{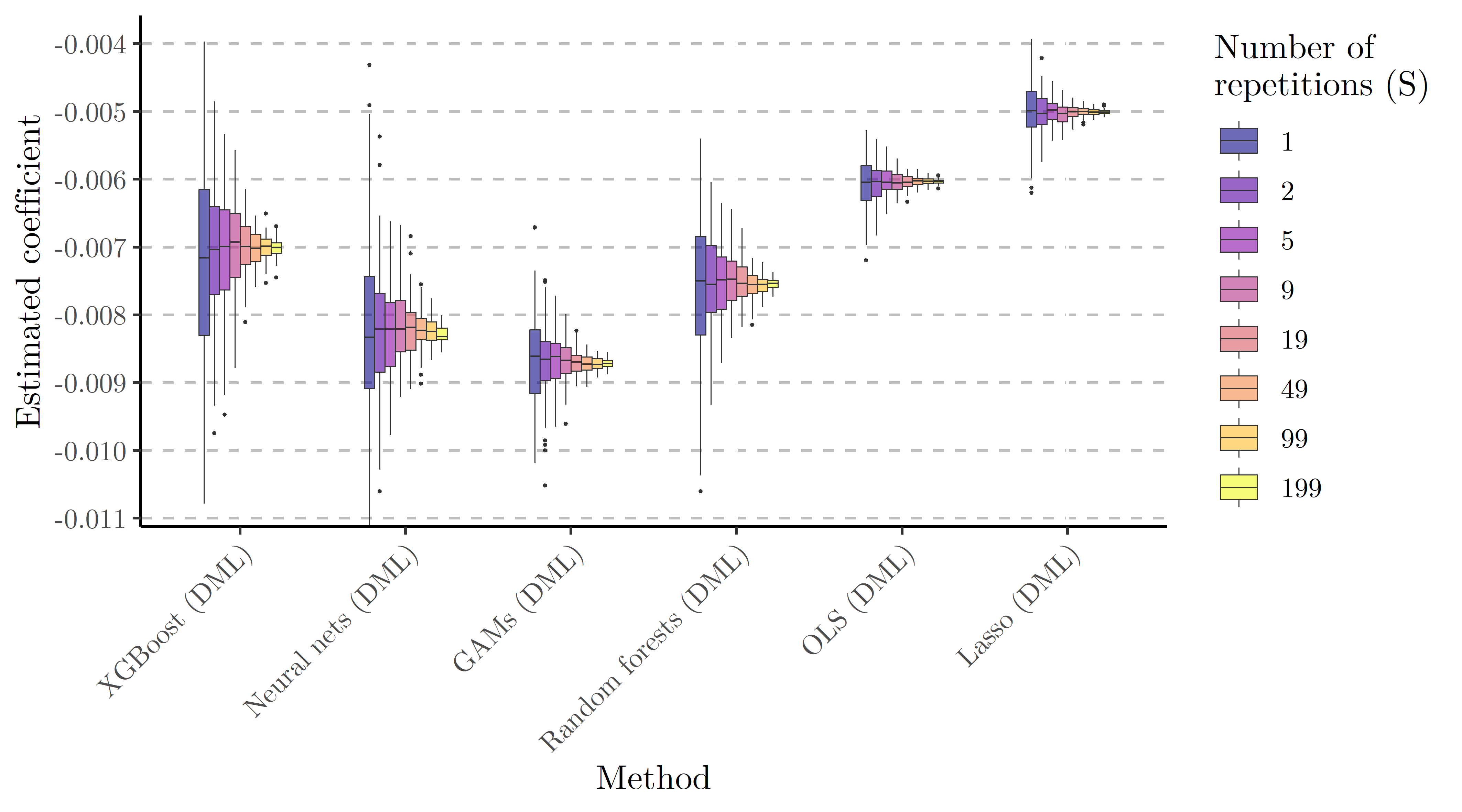}
    \caption{Varying the number of algorithm repetitions $S$ in DML when estimating the effect of air pollution on house prices. Each boxplot shows the distribution of estimated coefficients across 100 repeated runs when using DML with $S$ algorithm repetitions.}
    \label{fig:bysplits_app}
\end{figure}

The authors assume that some covariates influence both air pollution and house prices, while others only affect house prices (Figure \ref{fig:dag_appl}). Additionally, they argue for nonlinear relationships between housing attributes and house prices, thus transforming some of the variables accordingly. In their preferred specification (Equation \ref{eq:eq_harrison}), $mv$ is the median value of homes in a specific tract, $nox$ (concentration of nitrogen oxides\footnote{measured in parts per hundred million (pphm)}) is the measure of air pollution, and $\beta$ is the coefficient of interest. The other variables refer to the covariates in Figure \ref{fig:dag_appl} and are included with varying functional forms. The authors give theoretical justification for some of these functional forms, but choose others simply because they provide good fit. Given the large number of covariates and the authors' emphasis on the importance of specifying the correct functional forms, this application is an ideal example to demonstrate the strengths and flexibility of DML, which offers a data-driven way to flexibly learn functional forms and appropriately adjust for the covariates.
Variable descriptions are available in Table IV of \citet{harrison_hedonic_1978}, which we replicate in our Appendix (Table \ref{tab:appl_vardesc_appx}), where we also provide descriptive statistics (Table \ref{airpoll_descript}).
\begin{gather} \label{eq:eq_harrison}
\begin{aligned}
log(mv) = \alpha + \beta nox^2 &+ \gamma_1rm^2 + \gamma_2age + \gamma_3log(dis) + \gamma_4log(rad) + \gamma_5tax \\
&+ \gamma_6ptratio + \gamma_7(B-0.63)^2 + \gamma_8log(lstat) + \gamma_{9}crim \\
&+ \gamma_{10}zn + \gamma_{11}indus + \gamma_{12}chas  + \epsilon
\end{aligned}
\end{gather}
When we first applied DML with 5-fold cross-fitting to this dataset, we observed that repeated runs of the standard algorithm led to substantially different coefficient estimates, especially for the most flexible DML methods. Since this is likely a consequence of the random sample splitting in the first step, we systematically experimented with different numbers of repetitions $S$ to be more robust to this kind of randomness. For each number of repetitions, we run the full algorithm (including $S$ repetitions) 100 times and report the distribution of median estimates for each algorithm. For example, for the number of repetitions $S=5$, we estimate the model 5 times, compute the median estimate, and repeat this procedure for 100 times. The results indicate a clear pattern: the larger the number of repetitions $S$, the more stable the median coefficient estimates (Figure \ref{fig:bysplits_app}). This pattern is considerably more pronounced in this real-world dataset compared to our previously simulated data (Figure \ref{fig:bysplits}). One explanation is that not all variables in this real-world data neatly follow a symmetric distribution like the normal distribution in our simulations. As a consequence, the random sample splitting can lead to very heterogeneous samples for different runs. However, this is in essence an empirical question and therefore researchers should try different numbers of sample splits for their applications and choose a number at which the results become stable. For example, since we observed stable results for relatively large numbers of repetitions in this application, we use  $S=199$ for our final estimation and report the median for the effect estimates and standard errors. 

Table \ref{tbl:dml_appl_res} presents the results from applying DML to this real-world dataset. In the first three columns, we report the method, the effect estimate and the standard error, respectively. Since the air pollution variable is specified as nonlinear by the authors ($nox^2$), the estimated effect depends on the air pollution level at which it is evaluated. Thus, in column four, we display the percentage effect of a one-unit change in air pollution ($nox$) evaluated at the mean level of $nox$ (5.547 pphm). 
Columns five and six contain the mean squared error (MSE) of the respective ML method when predicting treatment and outcome in the first stage of DML. Under the assumption that the causal structure is correctly specified, we can use this as an indicator for which effect estimate to trust most, as we previously discussed in Section \ref{sim_predictiveness}.

\begin{table}[ht]
\centering
\caption{Results for the effect of air pollution on housing prices}
\label{tbl:dml_appl_res}
\scriptsize
\begin{tabular}[t]{lccccc}
\toprule
\textbf{Method} & \textbf{Effect estimate} & \textbf{Std. error} & \textbf{Effect at mean (\%)} & \textbf{MSE(Y)} & \textbf{MSE(W)}\\
\midrule
OLS (H\&R) & -0.0064 & 0.0011 & -7.08 & - & -\\
Simple OLS & -0.0146 & 0.0011 & -16.17 & - & -\\
XGBoost (naive) &  -0.0137 & 0.0013 & -15.14 & - & -\\
%naive\_rf &  -0.0128 & 0.0012 & -14.19 & - & -\\
OLS (raw) &  -0.0058 & 0.0011 & -6.47 & - & -\\
OLS (flex) & -0.0071 & 0.0013 & -7.88 & - & -\\
OLS (DML, flex) & -0.0093 & 0.0006 & -10.30 & 0.5571 & 1437.38\\
OLS (DML, raw) & -0.0059 & 0.0012 & -6.58 & 0.0402 & 58.91\\
OLS (DML, H\&R) & -0.0064 & 0.0012 & -7.14 & 0.0375 & 54.88\\
GAMs (DML) & -0.0087 & 0.0015 & -9.63 & 0.0346 & 42.71\\
Neural nets (DML) & -0.0081 & 0.0016 & -9.00 & 0.0349 & 34.46\\
Lasso (DML, flex) & -0.0071 & 0.0015 & -7.86 & 0.0316 & 33.89\\
XGBoost (DML) & -0.0070 & 0.0019 & -7.73 & 0.0295 & 20.85\\
Random forests (DML) & -0.0075 & 0.0018 & -8.27 & 0.0266 & 19.03\\
\bottomrule
\multicolumn{6}{p{\textwidth}}{\rule{0pt}{1em}\textit{Note: }MSE: mean squared error. H\&R: covariate specification by \citet{harrison_hedonic_1978}. raw: only using untransformed variables. flex: including squares and first-order interactions of all variables.}\\
\end{tabular}
%}
\end{table}

The first row replicates the original regression model specification of \citet{harrison_hedonic_1978} with an OLS regression, resulting in an effect estimate of -0.0064. Row two shows the results when not adjusting for any covariates, which leads to an absolutely larger effect estimate (-0.0146). The naive XGBoost implementation leads to a similarly large estimate, suggesting a similar inability to adjust for confounding. OLS without any transformed variables leads to an effect estimate that is smaller in absolute terms than the authors' nonlinear specification. The more flexible specification ``OLS (flex)" - including squares and first-order interactions of all variables - leads to a larger estimate of -0.0071. For all of these methods we do not observe a predictive accuracy in the last two columns.
However, when we implement DML with different predictive algorithms, we observe the predictive accuracy in the first stage. The DML results are sorted in descending order by the MSE of the first-stage prediction.
As the first DML implementation, we use an OLS regression in the first stage, while including squares and interactions of all variables. The results illustrate the problems with this approach: Since the number of parameters becomes large relative to the number of observations in each sample, the regression overfits significantly, as evident in the extremely poor predictive accuracy of the first stage. Also, the estimated effect is relatively large, in addition to varying extremely across repetitions of the algorithm. 
The next two methods again use OLS regression in DML, but now either use all variables linearly (``OLS (DML, raw)") or use the specification by \citet{harrison_hedonic_1978} (``OLS (DML, H\&R)"). This should give results similar to using the same specifications directly within an OLS regression. Within DML, however, we can also observe the predictive accuracy. The effect estimates are indeed very similar, and in addition, the MSEs for outcome and treatment indicate the merit of the authors' nonlinear specification: Their specification leads to a higher predictive accuracy compared to the purely linear model. 
The final five rows show results from using using DML with different flexible predictive algorithms. All flexible ML methods perform better than the OLS regression in the prediction tasks. Under the assumption that the identification is valid, we can thus expect the effect estimates of the flexible methods to be more precise than those of the OLS regression. There is also some variance in the flexible DML estimates, but they all agree in being more negative than the coefficient estimated in the authors' specification, suggesting that the authors in the original publication somewhat underestimate the true effect of air pollution. 
It is also worth discussing the relationship between the estimated effect and the directions of the confounding influence. The largest estimate in absolute terms comes from not adjusting at all  (``Simple OLS"), while the smallest estimate comes from adjusting only linearly (``OLS (raw)"). The more flexible and more credible specifications result in estimates significantly smaller than the naive approaches, but larger than the ones from linear adjustment. We can explain this observation by different confounders biasing the estimated effect in distinct directions. The linear specification might adjust well for confounders biasing the effect downwards, but poorly for confounders biasing the effect upwards. This heterogeneity in confounding direction is also supported by \citet{harrison_hedonic_1978}, who argue for a upward bias of some variables (e.g., distance to highways, distance to employment centers) and a downward bias of others (e.g., percentage lower status, crime rate, industry proportion) (see Figure \ref{fig:dag_appl_appx} in the Appendix).

\section{DML beyond the partially linear model} \label{sec:furth_settings}

So far, we have introduced and assessed DML for the partially linear model with one continuous treatment in a cross-sectional setting under the assumption of unconfoundedness. However, the general statistical theory translates to various other settings as well. In this section, we briefly mention further situations that differ in terms of the treatment, the data structure or the identification strategy.

\subsection{DML in the interactive model} \label{dml_interactive}

The partially linear regression (PLR) model assumes the treatment to have a homogeneous, additively separable effect on the outcome. We can apply DML with the PLR for both continuous and binary treatments.
However, for binary treatments, another version of DML allows for arbitrary effect heterogeneity by letting the treatment fully interact with the confounders \citep{chernozhukov_doubledebiased_2018}. Contrary to Equation \ref{eq:plry} for the PLR, the function $g_0()$ in Equation \ref{eq:inty} contains both the treatment and the confounders. 
\begin{gather} \label{eq:inty}
Y = g_0(W, \boldsymbol{X_c}) + V_y
\end{gather}
In this setting, we can use a slightly different estimator for the average treatment effect (ATE) without having to assume homogeneous treatment effects. In what follows, we briefly describe the intuition behind this estimator by first starting with two ``naive" ML estimators, which we then combine to form the estimator we can use for DML in the interactive model. 
The first ``naive" estimator for the ATE uses regression adjustment in the outcome model, that is:
\begin{gather} \label{eq:int_taunaive}
\displaystyle\hat\tau_{naive} = \frac1n \sum_{i=1}^n \Big(\hat\mu_{(1)}(X_i) - \hat\mu_{(0)}(X_i)\Big).
\end{gather}
Here, $\hat\mu_{(1)}(X_i)$ is the predicted outcome when we only use the treated observations for prediction; $\hat\mu_{(0)}(X_i)$ is the predicted outcome only using control observations. These predictions can come from any predictive method, including ML methods. However, as we discussed for the PLR, this estimator is biased if we do not specify the parametric model correctly, because we only consider the outcome model \citep{chernozhukov_doubledebiased_2018}.
The second ``naive" estimator in this setting relies on the estimated propensity score. When predicting a binary treatment from confounding variables, we can interpret the prediction as the (estimated) probability of receiving treatment (or the estimated ``propensity score", see \citealp{imbens_causal_2015}, Chapter 12). We can then use this estimated propensity score $\hat e(X_i)$ to weight observations by their probability of being treated, by which we make treatment and control group more comparable. This procedure is called ``inverse probability weighting" (IPW, or inverse propensity weighting, or Horvitz-Thompson estimator) and results in the estimator in Equation \ref{eq:int_tauipw}  \citep[see, e.g.,][]{hernan_causal_2020, rosenbaum_central_1983}. Here, we weight the treated units by the inverse of the estimated propensity score (first term), while we weight the control units (second term) by the inverse of one minus the estimated propensity score. 
\begin{gather} \label{eq:int_tauipw}
\displaystyle\hat\tau_{IPW} = \frac1n \sum_{i=1}^n \Big(\frac{W_i Y_i}{\hat e(X_i)} - \frac{(1 - W_i)Y_i}{1-\hat e(X_i)}\Big)
\end{gather}
We can estimate the propensity score for this method using any predictive method, including ML methods. However, the estimator can again be biased if we did not correctly specify the treatment/propensity model, because it does not consider the outcome model (\citealp{imbens_causal_2015}, Chapter 12).

Nevertheless, similar to the PLR, we can construct an unbiased estimator considering both the treatment and the outcome model simultaneously in a ``doubly robust" way. One name for this estimator is ``Augmented Inverse Propensity Weighting" (AIPW), because it augments the IPW estimator with the predictions from the outcome model \citep{robins_estimation_1994, robins_semiparametric_1995}. Thus, we can understand the AIPW estimator as a combination of the naive estimator $\hat\tau_{naive}$ and the IPW estimator $\hat\tau_{IPW}$. The innovation of \citet{chernozhukov_doubledebiased_2018} is using ML methods for both the propensity and the outcome model in combination with cross-fitting, which allows for flexible weighting and adjustment. The intuition for the estimator in Equation \ref{eq:int_tauaipw} is very similar to the DML estimator in the PLR. 
\begin{gather} \label{eq:int_tauaipw}
\begin{split} 
\displaystyle \hat{\tau}_{AIPW} = \frac{1}{n} \sum_{i=1}^n & \Big(\overbrace{\hat\mu_{(1)}(X_i) - \hat\mu_{(0)}(X_i)}^{\text{Regression adjustment using ML estimators}} \\ &  + \frac{W_i}{\hat e(X_i)} \underbrace{(Y_i - \hat\mu_{(1)}(X_i))}_{\text{residual treated}} - \frac{1-W_i}{1-\hat e(X_i)}\underbrace{(Y_i - \hat\mu_{(0)}(X_i))}_{\text{residual control}}\Big)
\end{split}
\end{gather}
The first term is exactly the naive estimator from before. The second term is similar to the IPW estimator, but instead of weighting the outcome variable $Y_i$, we weight the residual of the outcome variable. The outcome residual is the observed outcome $Y_i$ minus the prediction in the outcome model for treated ($\hat\mu_{(1)}(X_i)$) or control ($\hat\mu_{(0)}(X_i)$) units, respectively. 
The general structure of the DML algorithm remains. The cross-fitting procedure is still necessary, but here, we need to train two outcome models and use them for prediction, one for the treated observations ($\hat\mu_{(1)}(X_i)$), one for the control observations ($\hat\mu_{(0)}(X_i)$). The treatment model is now the propensity model, so we should make sure that the predictive/ML method returns a probability measure. With these three models, we can estimate $\hat{\tau}_{AIPW}$ for each fold and average the estimates across the folds to obtain the final estimate \citep{chernozhukov_doubledebiased_2018}. 

Like other estimators relying on the estimated propensity score, this estimator is sensitive to propensity score estimates that are too close to 0 or 1. In these cases, weighting by an extremely small number can inflate the residual and thus lead to a disproportionate impact of some units. There are multiple approaches for dealing with this issue in traditional methods \citep[see, e.g., ][]{imbens_causal_2015}. \citet{chernozhukov_doubledebiased_2018} use \textit{trimming} in one of their applications, where they trim the propensity score at extreme values. For this, one defines a trimming threshold (e.g., .01), and excludes all observations with propensity scores smaller than that threshold or larger than one minus that threshold (e.g., $<.01$ or $>.99$). In our experiments with DML in this setting, we observed that these extreme values for the propensity score occur more often for ML methods compared to traditional methods. While this could be a consequence of overfitting in poorly trained models, it could also indicate violations of the overlap assumption in these applications (\citealp{imbens_causal_2015}, Chapter 14).

\subsection{DML for instrumental variables models} \label{dml_IV}

In some applications, the unconfoundedness assumption might not be plausible. However, there could be an instrumental variable which we consider exogenous after adjusting for observed covariates. In these cases, DML can make the conditional exogeneity assumption of the instrument more plausible by flexibly adjusting for the observed covariates. The respective DML procedure is very similar to the unconfoundedness setting, though it now involves an additional prediction model for the instrumental variable \citep{chernozhukov_doubledebiased_2018}. For example, in the partially linear instrumental variables model, we make predictions and compute residuals for the treatment, outcome, \textit{and} instrument. Then, instead of using the residuals in a linear regression, we plug them into the standard IV estimator in place of the respective variables. For a binary treatment and a binary instrument, \citet{chernozhukov_doubledebiased_2018} demonstrate the estimation of local average treatment effects (LATE, \citet{imbens_identification_1994}), which allows the instrument to interact arbitrarily with the covariates.

\subsection{DML in further settings} \label{dml_further}

Here, we briefly mention other settings in which researchers might want to apply DML. 
First, like \citet{chernozhukov_doubledebiased_2018} in their first application, we can also use DML in experimental settings. There, the goal is not necessarily to adjust for observed confounding, which we can largely rule out if the random assignment was properly executed. Nevertheless, adjusting flexibly for covariates might help to increase the statistical precision of the estimates. Due to random assignment, we typically know the treatment or propensity model and do not need to estimate it, but using ML in the outcome model could be advantageous. 
Second, in some cases, we are interested in the effects of more than one treatment variable. If the effects of multiple variables are causally identified, we can simply compute the residual for each treatment and regress the outcome residual on all of the multiple treatment residuals \citep{chernozhukov_doubledebiased_2018}. 
Similarly, DML translates to settings with multi-valued treatments or multiple binary treatment versions \citep[see, e.g.,][]{knaus_double_2022}. 
Fourth, there is currently a lack of clear guidance about whether or how DML extends to settings with panel or longitudinal data. \citet{chernozhukov_doubledebiased_2018} focus on a cross-sectional setting, and adaptations to panel settings do not seem trivial for two reasons: First, the time dimension complicates the cross-fitting procedure; second, the elimination of unobserved heterogeneity is not straightforward in complex nonlinear models.

\section{Discussion} \label{sec:discussion}

Our simulations and the application have demonstrated the capability of DML to adjust for various forms of confounding influences, as long as flexible ML methods are used (e.g., boosted trees, neural nets, or random forests), and the sample size is reasonably large. Also, a sound choice of parameters within the DML algorithm can improve the accuracy (e.g., number of folds) and robustness (e.g., number of repetitions) of the estimates. Furthermore, we have illustrated that accuracy in predicting treatment and outcome is indicative of accuracy of the effect estimation. 
Additionally, we have highlighted that assumptions about causal structure are still required, as DML cannot account for unobserved confounding or recognize a bad control. 
Based on these findings, in what follows we derive recommendations for applied researchers about when and how to apply DML (see Table \ref{tab:recommend}).

\subsection{\textit{When} should we (not) use DML?}

First, while we have shown DML to be a useful estimation method, it is \textit{not} a novel strategy for causal \textit{identification}. That is, if unobserved confounding is a problem in a specific application, DML is not a cure for it. Instead, DML can serve as a way of estimating the effect from the data \textit{after} one has decided for a plausible identification strategy. Then, DML can estimate the effect more flexibly than traditional methods by modeling potentially complex functional forms in a data-driven way. In our study, we have focused on using DML for settings with the identification strategy ``adjusting for observed confounders". However, DML is also applicable for strategies like randomized experiments (where it can potentially increase precision) and instrumental variables, with researchers continuously exploring extensions to further strategies (e.g., difference-in-differences \citep{chang_doubledebiased_2020}). %, see Table \ref{tab:extns}). 
For strategies that depend on adjusting for covariates well, DML may provide a solution that does not rely on correctly pre-specified functional forms. In summary, DML does not help to relax identification assumptions, but it can relax estimation assumptions. For any application, we should first use theory and domain knowledge to outline a plausible causal structure (which we can represent, e.g., with a DAG). If one of the mentioned identification strategies is feasible in this causal structure, researchers can use DML to estimate the effect without making strong parametric assumptions. 
Second, if researchers find themselves in a situation with cross-sectional data, where identification is possible, and there is no strong theoretical basis for choosing specific functional forms, DML offers a promising alternative to arbitrarily specifying a parametric model. Even in cases where theory suggests functional relationships for some confounders, we recommend using DML as it might help to assess the robustness of the specification to different functional forms learned from the data. 
Third, DML's ability to flexibly adjust for covariates is especially pronounced for sample sizes significantly larger than the number of important raw covariates. The method's benefit over parametric methods will likely be minor for very small samples, so we recommend using DML when the number of observations significantly exceeds the number of important raw covariates. 
Finally, for the moment, we can give no recommendation for the use of DML with panel data. The original paper of \citet{chernozhukov_doubledebiased_2018} focuses on establishing DML for settings with cross-sectional data. While there is gradual process for understanding DML in panel settings and the value it can add there, at the time of writing, there are no theoretical guarantees comparable to the cross sectional setting and there is no clear practical guidance available.

\subsection{\textit{How} should we use DML?}

Once we have decided that using DML is appropriate in our application, we still have a plethora of decisions to make in the process of applying DML. 
The first decision is which variables to include in the  estimation. This is related to the question of identification, but goes further. For identification, we must adjust for all confounders ($X_c$), meaning variables influencing both the treatment (or instrument in instrumental variables settings) and the outcome. Secondly, we must not include any bad controls such as colliders ($X_{coll}$) \citep[see, e.g.,][]{angrist_mostly_2009, cinelli_crash_2022}. These two criteria ensure identification, but other types of variables also can influence the precision of the estimates. We should include variables only influencing the outcome ($X_p$), should not include variables only influencing the treatment (or instrument) ($X_z$), and can, but do not have to, include noise variables ($E$). We recommend using DAGs to systematically consider to which of these categories each covariate belongs. 

The second decision is which ML algorithm to use within DML. 
The literature overview in Section \ref{sec:literature} has shown that most published applications of DML rely on lasso regressions. In contrast, based on our simulation analyses, we caution against the use of lasso because DML using lasso with raw covariates produces biased estimates as soon as the confounding is not only linear. We rather recommend flexible methods with the ability to fit any functional forms, like random forests, boosted trees, or neural networks. Ideally, researchers should tune these methods to achieve high quality predictions and avoid overfitting. If the computational resources allow it, running DML with different ML algorithms can help to assess the robustness of the estimates. Also, estimating a traditional parametric model is a helpful reference point. If all methods agree, we can have confidence that the estimates are robust to different functional forms. If the DML methods agree, but are very different from the parametric model, we might want to question the plausibility of the parametric model. If different DML methods lead to very different estimates, we can proceed in three different ways: (1) We could return to the causal structure and question whether our identification strategy is valid. If we are confident that it is, we could (2) use the results as as form of sensitivity analysis and report the results as a range of plausible estimates. (3) If we want to obtain a most plausible point estimate, we could use the predictive accuracy of each ML algorithm in the first stage to determine which parameter estimate we trust most. If we have entered the correct variables, methods performing better at predicting treatment and outcome should also adjust more completely for confounding and thus deliver more accurate causal estimates. If we observe different algorithms performing best for outcome and treatment, respectively, it is also possible to use the best-performing method for the respective prediction task. If we had to pick the one single ML method that performs best, we would recommend the use of boosted trees in the XGBoost implementation. The reason is that XGBoost performs very well across a very broad range of settings, and we could rarely identify any setting in which XGBoost is clearly dominated by a different method. 

Third, we must decide about the number of folds $K$ into which we split the dataset in DML. One important factor for this decision is the sample size. For small samples, we should choose moderate to large numbers of folds (5-10 folds), because they provide more observations to the prediction task, which increases the flexibility of the ML methods. However, $K$ should not be so large that very few observations remain in the fold we use for estimating the effect. In larger samples, the choice of $K$ is less influential. Since large values for $K$ quickly increase computation times, a choice of $K=2$ is reasonable for very large samples. However, we still recommend a moderate number of folds (e.g., $K=5$) if computational resources allow it. 

Finally, we have to choose the number of DML repetitions $S$ for achieving robust estimates. We recommend treating this as an empirical question: For any specific application, researchers can run DML with different $S$ and observe whether the estimates of repeated runs are stable. As we illustrated in Section \ref{sec:application}, we should finally choose the smallest $S$ with stable estimates because larger numbers of $S$ significantly increase computation times. 

\def\arraystretch{1.3}%  1 is the default, change whatever you need

\begin{table}[ht]
\scriptsize
\caption{Problems and recommendations for the application of DML}
\resizebox{\columnwidth}{!}{
\begin{tabular}{>{\raggedright}p{.3\textwidth} p{.7\textwidth}}
\toprule
\textbf{Decision problem}    & \textbf{Recommended decision}                                                                                                                    \\ \midrule
Is the effect causally identified?    & Determine identification independently of DML. If the effect is identified based on observables,  use DML to flexibly adjust for covariates.                             \\
Which variables should we include?  & Confounders, outcome influencers, (noise variables), no instruments, no bad controls.                             \\
Which ML algorithm should we use?   & Use flexible ML methods like random forests, XGBoost, or neural networks, ideally with some parameter tuning. Choose between ML algorithms based on the predictive accuracy in the first stage.            \\
How many folds $K$ should we use for cross-fitting?   & Small samples: $K = 5 - 10$; larger samples: $K = 2 - 5$.                             \\
How many repetitions $S$ should we use?   &  Test different sizes of $S$, choose the lowest $S$ that still leads to stable estimates.                            \\
What if we have a very small sample size (e.g., $n < 100$)? & Be aware of DML's limited flexibility in small samples. Run a plausible parametric model in addition to DML and compare the estimates. \\
What if we are confident in our parametric model? & You can still use DML as a robustness check to rule out functional form misspecifications. \\
 \bottomrule
\end{tabular}
\label{tab:recommend}
}
\end{table}

\subsection{Limitations of our study}

Our study evaluates the DML method in a variety of settings, but there are interesting settings and implementations we did not consider. First, in our simulations, we focus on the partially linear model under unconfoundedness with a continuous and homogeneous treatment. It would be interesting to see how our conclusions translate to settings with binary treatments, treatment heterogeneity and unobserved confounding with instrumental variables. 
Second, \citet{chernozhukov_doubledebiased_2018} describe two different score functions for employing DML in the partially linear model. We only use the ``partialling-out" score function and do not contrast it with the ``IV-type" score function, since we did not see significant differences in a few initial simulations. 
Third, one could compare additional ML algorithms within DML, implement DML with distinct ML methods for the treatment and the outcome model, respectively, and compare DML to further traditional or ML-based statistical methods. 
Fourth, in most simulations, we are changing one or two factors of the DGP while holding all others constant. This is a practical strategy, but might not uncover all nuances of how different characteristics of the data are interrelated. 
Lastly, we assess DML in a cross-sectional setting without a time dimension. However, many disciplines regularly deal with repeated observations over time in panel or longitudinal data. In these settings,  methods like fixed effects estimation can eliminate the influence of time-constant, unobserved variables (\citealp{wooldridge_econometric_2010}, Chapter 10). The feasibility of DML in these settings is not yet well understood, since the time-dimension complicates the cross-fitting procedure and nonlinear relationships might hinder the elimination of the unobserved effects. Exploring if and how we can adapt DML to allow estimation under weaker assumptions in these settings is an important avenue for further research.

\section*{Acknowledgements}
Funded by the Deutsche Forschungsgemeinschaft (DFG, German Research Foundation) under Germany’s Excellence Strategy – EXC number 2064/1 – Project number 390727645. 
The authors acknowledge support by the state of Baden-Württemberg through bwHPC. Philipp Berens acknowledges support by the Hertie Foundation.

\clearpage
\bibliography{export_bib}
\bibliographystyle{apalike}
\clearpage

\appendix
\section{Appendix} \label{appendix}

\subsection{Literature selection} \label{appx:lit_selection}

For the literature review, we identified 46 papers published up to March 2023, which cite the initial proposal of DML in \citet{chernozhukov_doubledebiased_2018}, and apply the method to real-world data. From these, we excluded papers that contained too little information about the implementation of DML or the data used, either because it was not reported or because we could not access the full paper. At a minimum, the authors had to report which ML algorithm they used and their sample size. We also excluded papers that use the double selection method by \citet{belloni_high-dimensional_2014}, but call it DML. If the study contained multiple different specifications, we recorded the lowest sample size $n$ and the largest number of raw covariates $p$ to get a conservative estimate of the dimensionality. If the authors did not state number of covariates explicitly, we counted all covariates mentioned in the text. Typically, the authors transformed categorical covariates into dummy variables. We counted $p$ after that transformation; in cases where the number of levels of a categorical variable is unclear, we guessed it. The covariate dimension $p$ always refers to the raw covariates, before (some of) the authors added nonlinearities through polynomials, interactions or other transformations.

\def\arraystretch{1.1}

\begin{landscape}\begin{table}[!ht]

\caption{\label{tab:appl_appx}Published papers applying DML}
\centering
\resizebox{\linewidth}{!}{
\fontsize{6}{7.5}\selectfont
\begin{tabular}[t]{>{\raggedright\arraybackslash}p{.02\columnwidth}>{\raggedright\arraybackslash}p{.13\columnwidth}>{\raggedright\arraybackslash}p{.55\columnwidth}>{\raggedright\arraybackslash}p{.15\columnwidth}>{\raggedright\arraybackslash}p{.15\columnwidth}}
\toprule
\textbf{Year} & \textbf{Author} & \textbf{Title} & \textbf{Journal} & \textbf{Discipline}\\
\midrule
2018 & \citeauthor{chernozhukov_doubledebiased_2018} & Double/debiased machine learning for treatment and structural parameters & The Econometrics Journal & Statistics / Econometrics\\
2019 & \citeauthor{mcconnell_estimating_2019} & Estimating treatment effects with machine learning & Health Services Research & Healthcare / Medicine\\
2020 & \citeauthor{chang_doubledebiased_2020} & Double/debiased machine learning for difference-in-differences models & The Econometrics Journal & Statistics / Econometrics\\
2020 & \citeauthor{dube_monopsony_2020} & Monopsony in Online Labor Markets & American Economic Review: Insights & Economics\\
2020 & \citeauthor{hansen_effectiveness_2020} & The effectiveness of fiscal institutions: International financial flogging or domestic constraint? & European Journal of Political Economy & Political Science\\
2020 & \citeauthor{holtz_interdependence_2020} & Interdependence and the cost of uncoordinated responses to COVID-19 & Proceedings of the National Academy of Sciences & Economics\\
2020 & \citeauthor{yang_double_2020} & Double machine learning with gradient boosting and its application to the Big N audit quality effect & Journal of Econometrics & Economics\\
2021 & \citeauthor{azoulay_long-term_2021} & Long-term effects from early exposure to research: Evidence from the NIH "Yellow Berets" & Research Policy & Healthcare / Medicine\\
2021 & \citeauthor{chan_behind_2022} & Behind the screen: Understanding national support for a foreign investment screening mechanism in the European Union & The Review of International Organizations & Political Science\\
2021 & \citeauthor{chen_debiaseddouble_2021} & Debiased/Double Machine Learning for Instrumental Variable Quantile Regressions & Econometrics & Statistics / Econometrics\\
2021 & \citeauthor{chernozhukov_causal_2021} & Causal impact of masks, policies, behavior on early covid-19 pandemic in the U.S. & Journal of Econometrics & Economics\\
2021 & \citeauthor{knaus_double_2021} & A double machine learning approach to estimate the effects of musical practice on student's skills & Journal of the Royal Statistical Society: Series A (Statistics in Society) & Economics\\
2021 & \citeauthor{kuppelwieser_liquidity_2021} & Liquidity costs on intraday power markets: Continuous trading versus auctions & Energy Policy & Economics\\
2021 & \citeauthor{liu_doubledebiased_2021} & Double/debiased machine learning for logistic partially linear model & The Econometrics Journal & Statistics / Econometrics\\
2021 & \citeauthor{parpouchi_association_2021} & The association between experiencing homelessness in childhood or youth and adult housing stability in Housing First & BMC Psychiatry & Healthcare / Medicine\\
2021 & \citeauthor{semenova_debiased_2021} & Debiased machine learning of conditional average treatment effects and other causal functions & The Econometrics Journal & Statistics / Econometrics\\
2021 & \citeauthor{skoufias_child_2021} & Child stature, maternal education, and early childhood development in Nigeria & PLOS ONE & Interdisciplinary\\
2021 & \citeauthor{yamane_is_2021} & Is the younger generation a driving force toward achieving the sustainable development goals? Survey experiments & Journal of Cleaner Production & Sociology\\
2022 & \citeauthor{alley_pricing_2022} & Pricing for Heterogeneous Products: Analytics for Ticket Reselling & Manufacturing \& Service Operations Management & Economics\\
2022 & \citeauthor{bilancini_prosocial_2022} & Prosocial behavior in emergencies: Evidence from blood donors recruitment and retention during the COVID-19 pandemic & Social Science \& Medicine & Interdisciplinary\\
2022 & \citeauthor{bodory_evaluating_2022} & Evaluating (weighted) dynamic treatment effects by double machine learning & The Econometrics Journal & Statistics / Econometrics\\
2022 & \citeauthor{cardenas_youth_2022} & Youth well-being predicts later academic success & Scientific Reports & Interdisciplinary\\
2022 & \citeauthor{chiang_multiway_2022} & Multiway Cluster Robust Double/Debiased Machine Learning & Journal of Business \& Economic Statistics & Statistics / Econometrics\\
2022 & \citeauthor{decarolis_corruption_2022} & Corruption red flags in public procurement: New evidence from Italian calls for tenders & EPJ Data Science & Economics\\
2022 & \citeauthor{ellickson_estimating_2022} & Estimating Marketing Component Effects: Double Machine Learning from Targeted Digital Promotions & Marketing Science & Economics\\
2022 & \citeauthor{farbmacher_causal_2022} & Causal mediation analysis with double machine learning & The Econometrics Journal & Statistics / Econometrics\\
2022 & \citeauthor{goller_analysing_2023} & Analysing a built-in advantage in asymmetric darts contests using causal machine learning & Annals of Operations Research & Sports Research\\
2022 & \citeauthor{gordon_close_2022} & Close Enough? A Large-Scale Exploration of Non-Experimental Approaches to Advertising Measurement & Marketing Science & Economics\\
2022 & \citeauthor{huber_business_2022} & Business analytics meets artificial intelligence: Assessing the demand effects of discounts on Swiss train tickets & Transportation Research Part B: Methodological & Economics\\
2022 & \citeauthor{knaus_double_2022} & Double machine learning-based programme evaluation under unconfoundedness & The Econometrics Journal & Statistics / Econometrics\\
2022 & \citeauthor{loiseau_external_2022} & External control arm analysis: An evaluation of propensity score approaches, G-computation, and doubly debiased machine learning & BMC Medical Research Methodology & Healthcare / Medicine\\
2022 & \citeauthor{lundberg_gap-closing_2022} & The Gap-Closing Estimand: A Causal Approach to Study Interventions That Close Disparities Across Social Categories & Sociological Methods \& Research & Sociology\\
2022 & \citeauthor{qiu_statistical_2022} & Statistical and machine learning methods for evaluating trends in air quality under changing meteorological conditions & Atmospheric Chemistry and Physics & Geoscience\\
2022 & \citeauthor{vansteelandt_assumption-lean_2022} & Assumption-Lean Cox Regression & Journal of the American Statistical Association & Statistics / Econometrics\\
2023 & \citeauthor{dickson_adherence_2023} & Adherence, Persistence, Readmissions, and Costs in Medicaid Members with Schizophrenia or Schizoaffective Disorder Initiating Paliperidone Palmitate Versus Switching Oral Antipsychotics: A Real-World Retrospective Investigation & Advances in Therapy & Healthcare / Medicine\\
2023 & \citeauthor{felderer_using_2023} & Using Double Machine Learning to Understand Nonresponse in the Recruitment of a Mixed-Mode Online Panel & Social Science Computer Review & Sociology\\
\bottomrule
\end{tabular}}
\end{table}
\end{landscape}

\FloatBarrier

\subsection{Notes for replication}

In this appendix section, we note additional settings in our code not mentioned in the main text that should ensure replicability of our results. 

For all computations, we use the random seed 42 and compute in parallel on 50 cores on the HPC bwUniCluster. For parallel computing, we use the package \textit{foreach} \citep{daniel_foreach_2022} in combination with the \textit{doParallel} backend \citep{daniel_doparallel_2022} and rely on the package \textit{doRNG} \citep{gaujoux_dorng_2023} for replicable random number generation on multiple cores.

\subsection{Figures} \label{app_figures}

\subsubsection{Comparing baseline results for different numbers of simulation iterations \label{appx_nsim}}

\begin{figure}[!ht]
    \centering
    \includegraphics[width=\textwidth]{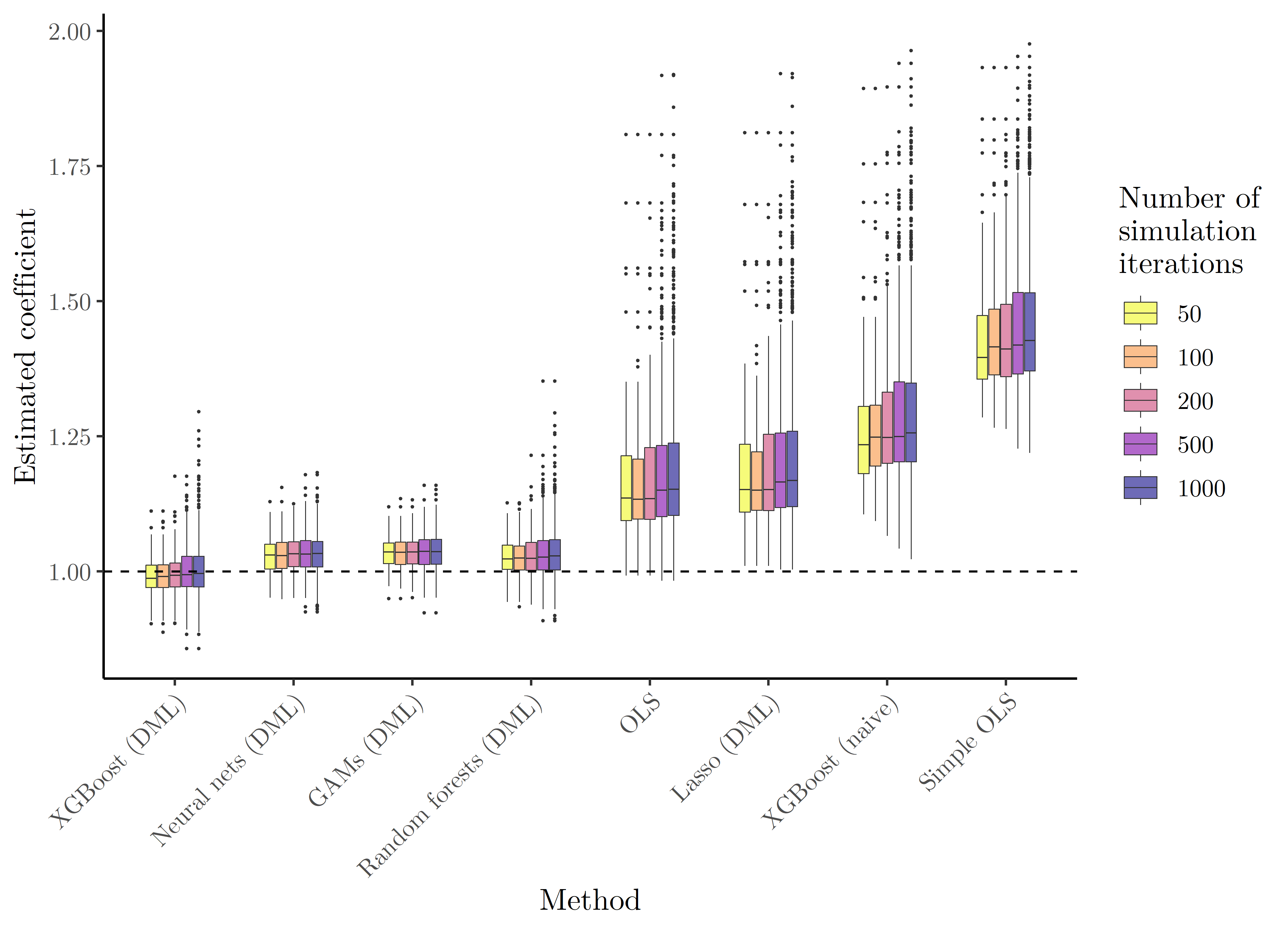}
    \caption{Comparing baseline results for different numbers of simulation iterations}
    \label{fig:bynsim}
\end{figure}

Figure \ref{fig:bynsim} shows results for the ``baseline" simulation setting for 50 to 1,000 simulation iterations. The overall results are very stable. As expected, adding more iterations creates more outliers, especially for the methods not adjusting well for confounding. Based on these results, we choose 100 simulation iterations for the remainder of our simulations, as the resulting distribution seems fairly representative, while using computing resources efficiently. 

\FloatBarrier

\subsection{Application notes}

\begin{figure}[ht]
\begin{minipage}[top]{\textwidth}
%\justifying
    \begin{center}
      \begin{tikzpicture}
      \small
          \node (w) at (0,0) [label=below:air pollution,point];
          \node (y) at (8,0) [label=below:house price,point];
        %   \node[align = center] (xc) at (3,3) {industry proportion \\ distance employment \\  zoned large lots \\ highways \\  crime rate \\ river};
          \node[align = center, fill = white] (xctext) at (4,1.7) {+ \% lower status $-$\\ ($-$/?) river +   \\ + highways + \\ (+) crime rate $-$ \\  (+) \% industry $-$ \\ (?) zoned large lots +  \\ $-$ distance employment $-$ };
          \node (xc) at (4, 2.5) {};
          \node[align = center] (xp) at (8,2) { pupil-teacher ratio $-$ \\ property-tax rate $-$ \\  \% blacks $U$ \\   rooms + \\ age ($-$)};
    
          \begin{pgfonlayer}{bg}  
              \path (w) edge (y);
              \path (xc) edge (y);
              \path (xc) edge (w);   
              \path (xp) edge (y);
          \end{pgfonlayer}
      \end{tikzpicture}
    \end{center}
\end{minipage}
\caption{\small \label{fig:dag_appl_appx}Causal structure argued for in \citet{harrison_hedonic_1978}, including hypothesized effect signs.  $+$/$-$ are explicitly stated, ($+$)/($-$) are implicitly inferred. $U$ refers to a U-shaped parabolic relationship.}
\end{figure}
\vspace{-2ex}

\newpage

\newdimen\NetTableWidth
\noindent
  \NetTableWidth=\dimexpr
    \linewidth
    - 8\tabcolsep
    - 5\arrayrulewidth % if package array is loaded
  \relax
  
\def\arraystretch{1.2}

%\begin{table}[ht]
\begin{spacing}{1.1}
\begin{longtable}{%
    p{.15\NetTableWidth}%
    p{.65\NetTableWidth}%
    p{.2\NetTableWidth}%
    }
\caption{\label{tab:appl_vardesc_appx}Variable descriptions (replication of Table IV in \citet{harrison_hedonic_1978})}
 \\
 \toprule
  \textbf{Variable} & \textbf{Definition} & \textbf{Source} \\
 \midrule
$MV$    & Median value of owner-occupied homes.    & 1970 U. S. Census \\
 $RM$    & Average number of rooms in owner units. $RM$ represents spaciousness and, in a certain sense, quantity of housing. It should be positively related to housing value. The $RM^2$ form was found to provide a better
fit than either the linear or logarithmic forms.   & 1970 U. S. Census\\
 $AGE$    & Proportion of owner units built prior to 1940. Unit age is generally related to structure quality. & 1970 U. S. Census\\
 $B$    & Black proportion of population. At low to moderate levels of $B$, an increase in $B$ should have a negative influence on housing value if Blacks are regarded as undesirable neighbors by Whites. However, market discrimination means that housing values are higher at very high levels of $B$. One expects, therefore, a parabolic relationship between proportion Black in a neighborhood and housing values. 
  \footnote{Comment by JF: This is a rather controversial variable (see, e.g., https://medium.com/@docintangible/racist-data-destruction-113e3eff54a8). With the authors' functional form, it is in a way modeling systemic racism. This is problematic if the model were to be used to set prices for homes based on this variable. However, since the original authors (and we) are interested in the effect of air pollution on housing prices, this variable is only included to adjust for how past racism might bias this effect. In this sense, it describes and adjusts for the discrimination \textit{at that time}; we do not suggest to use it to predict prices, and its use is no endorsement of systemic racism.
  An additional problem is that the available data does not contain the ``black" variable in its raw form, that is, the black proportion of the population. It only contains the parabolic transformation, which is a non-invertible function, so we cannot transform it back to the original raw values. One can argue against this functional form, but we expect using the variable with a wrong functional form (which the ML methods might even improve) is still more informative than not using it at all. Because of this, we use the transformed variable (``B\_trans") throughout our analysis.}
  & 1970 U. S. Census 
 \\
 $LSTAT$ & Proportion of population of that is lower status = $\frac{1}{2}$ (proportion of adults without some high school education and proportion of male workers classified as laborers). The logarithmic specification implies that socioeconomic status distinctions mean more in the upper brackets of society than in lower classes.  & 1970 U. S. Census  \\
 $CRIM$ & Crime rate by town. Since $CRIM$ gauges the threat to well-being that households perceive in various neighborhoods of the Boston metropolitan area (assuming that crime rates are generally proportional to people’s perceptions of danger) it should have a negative effect on housing values. & FBI (1970) \\
 $ZN$ & Proportion of a town’s residential land zoned for lots greater than 25,000 square feet. Since such zoning restricts construction of small lot houses, we expect $ZN$ to be positively related to housing values. A positive coefficient may also arise because zoning proxies the exclusivity, social class, and outdoor amenities of a community.  & Metropolitan Area Planning Commission (1972)\\
 $INDUS$&   Proportion non-retail business acres per town. $INDUS$ serves as a proxy for the externalities associated with industry-noise, heavy traffic, and unpleasant visual effects, and thus should affect housing values negatively. & Vogt, Ivers and Associates [33]\\
$TAX$ & Full value property tax rate (\$/\$10,000). Measures the cost of public services in each community. Nominal tax rates were corrected by local assessment ratios to yield the full value tax rate for each town. Intratown differences in the assessment ratio were difficult to obtain and thus not used. The coefficient of this variable should be negative. & Massachusetts Taxpayers Foundation (1970)\\
    $PTRATIO$ & Pupil-Teacher ratio by town school district. Measures public sector benefits in each town. The relation of the pupil-teacher ratio to school quality is not entirely clear, although a low ratio should imply each student receives more individual attention. We expect the sign on $PTRATIO$ to be negative. & Massachusetts Dept. of Education (1971-1972)\\
    $CHAS$ & Charles River dummy: =1 if tract bounds the Charles River; =0 if otherwise. $CHAS$ captures the amenities of a riverside location and thus the coefficient should be positive.  & 1970 U. S. Census Tract maps\\
    $DIS$ & Weighted distances to five employment centers in the Boston region. According to traditional theories of urban land rent gradients, housing values should be higher near employment centers. $DIS$ is entered in logarithm form; the expected sign is negative. & Schanre [29] \\
    $RAD$ & Index of accessibility to radial highways. The highway access index was calculated on a town basis. Good road access variables are needed so that auto pollution variables do not capture the locational advantages of roadways. $RAD$ captures other sorts of locational advantages besides nearness to workplace. It is entered in logarithmic form; the expected sign is positive. & MIT Boston Project \\
    $NOX$ & Nitrogen oxide concentrations in pphm (annual average concentration in parts per hundred million). & TASSIM \\
    $PART$ & Particulate concentrations in mg/hcm$^3$ (annual average concentration in milligrams per hundred cubic meters) & TASSIM \\
\bottomrule
\end{longtable}
\end{spacing}

\begin{table}[ht]
\centering
%\resizebox{\linewidth}{!}{
\caption{Descriptive statistics}
\label{airpoll_descript}
\scriptsize
\begin{tabular}[t]{lccccccc}
\toprule
\textbf{Variable} & \textbf{Minimum} & \textbf{Q1} & \textbf{Median} & \textbf{Mean} & \textbf{Q3} & \textbf{Maximum} & \textbf{SD}\\
\midrule
medv & 5000 & 17025 & 21200 & 22532 & 25000 & 50000 & 9197\\
nox & 3.85 & 4.49 & 5.38 & 5.54 & 6.24 & 8.71 & 1.15\\
crim & 0.006 & 0.082 & 0.256 & 3.613 & 3.677 & 88.976 & 8.601\\
zn & 0.00 & 0.00 & 0.00 & 11.36 & 12.50 & 100.00 & 23.32\\
indus & 0.46 & 5.19 & 9.69 & 11.13 & 18.10 & 27.74 & 6.86\\
chas & 0.00 & 0.00 & 0.00 & 0.07 & 0.00 & 1.00 & 0.25\\
rm & 3.56 & 5.88 & 6.20 & 6.28 & 6.62 & 8.78 & 0.70\\
age & 2.90 & 45.02 & 77.50 & 68.57 & 94.07 & 100.00 & 28.14\\
dis & 1.12 & 2.10 & 3.20 & 3.79 & 5.18 & 12.12 & 2.10\\
rad & 1.00 & 4.000 & 5.00 & 9.54 & 24.00 & 24.00 & 8.70\\
tax & 187.0 & 279.00 & 330.0 & 408.2 & 666.0 & 711.0 & 168.5\\
ptratio & 12.60 & 17.40 & 19.05 & 18.45 & 20.20 & 22.00 & 2.16\\
B\_trans & 0.0003 & 0.3754 & 0.3914 & 0.3567 & 0.3962 & 0.3969 & 0.0913\\
lstat & 0.0173 & 0.0695 & 0.1136 & 0.1265 & 0.1696 & 0.3797 & 0.0714\\
\bottomrule
\end{tabular}%}
\end{table}

\FloatBarrier

\clearpage

\end{document}